\documentclass[10pt,journal,compsoc]{IEEEtran}

\usepackage{xspace}
\makeatletter
\DeclareRobustCommand\onedot{\futurelet\@let@token\@onedot}
\def\@onedot{\ifx\@let@token.\else.\null\fi\xspace}
\def\eg{\emph{e.g}\onedot} 
\def\ie{\emph{i.e}\onedot} 
 
\def\etc{\emph{etc}\onedot} 
 
\def\etal{\emph{et al}\onedot}
\makeatother

\usepackage{times}
\usepackage{epsfig}
\usepackage{float}
\usepackage{graphicx}
\usepackage{amsmath}
\usepackage{amssymb}
\usepackage{multirow}
\usepackage{algorithm}
\usepackage{hhline} 

\usepackage{color}
\usepackage{colortbl}

\usepackage[nocompress]{cite}
\usepackage{url}
\usepackage{xspace}
\usepackage{xcolor}
\usepackage{indentfirst}
\usepackage{ragged2e}
\usepackage{makecell}
\usepackage{diagbox}
\usepackage{bm}
\usepackage{booktabs} 
\usepackage[misc]{ifsym}
\definecolor{myblue}{RGB}{0, 150, 199}
\definecolor{mypink2}{RGB}{239, 71, 111}
\definecolor{mypink3}{RGB}{255,153,0}
\definecolor{myblue2}{RGB}{69, 123, 157}
\definecolor{myblue1}{RGB}{189, 227, 229}
\definecolor{mygray}{RGB}{208, 206, 206}

\usepackage{pifont}% http://ctan.org/pkg/pifont
\usepackage[T1]{fontenc}
\usepackage{multirow}
\usepackage[rightcaption]{sidecap}
\usepackage{makecell}
\usepackage{xspace}
\usepackage{enumitem}
\usepackage{url}
\usepackage{enumerate}
\usepackage{wrapfig}
\usepackage[pagebackref=false,breaklinks=true,bookmarks=false,citecolor={black}, colorlinks=true,linkcolor={black}]{hyperref}

\usepackage{amsmath, amssymb} % define this before the line numbering.
\usepackage{booktabs}
\usepackage{bm}
\usepackage{textcomp}
\usepackage{gensymb}
\usepackage{pifont}% http://ctan.org/pkg/pifont
\usepackage[T1]{fontenc}
\usepackage{multirow}
\usepackage[rightcaption]{sidecap}
\usepackage{makecell}
\usepackage{xspace}
\usepackage{enumitem}
\usepackage{url}
\usepackage{enumerate}
\usepackage{wrapfig}
\usepackage[noend]{algpseudocode} 
\usepackage[colorinlistoftodos]{todonotes}
\usepackage{amsmath,amssymb} % define this before the line numbering.

\makeatletter
\DeclareRobustCommand\onedot{\futurelet\@let@token\@onedot}
\def\@onedot{\ifx\@let@token.\else.\null\fi\xspace}
\def\eg{\emph{e.g}\onedot,~} 
\def\ie{\emph{i.e}\onedot,~} 
 
\def\etc{\emph{etc}\onedot} 
 
\def\etal{\emph{et al}\onedot}

\makeatother

\newcommand{\cmark}{\ding{51}}%
\newcommand{\xmark}{\ding{55}}%

\newcommand{\PreserveBackslash}[1]{\let\temp=\\#1\let\\=\temp}
\newcolumntype{C}[1]{>{\PreserveBackslash\centering}p{#1}}
\newcolumntype{R}[1]{>{\PreserveBackslash\raggedleft}p{#1}}
\newcolumntype{L}[1]{>{\PreserveBackslash\raggedright}p{#1}}
\begin{document}
%
% paper title
\title{Attribute Descent: Simulating Object-Centric Datasets on the Content Level and Beyond}

\author{Yue Yao,
    Liang Zheng,
    Xiaodong Yang,
    Milind Napthade,
    and Tom Gedeon% <-this % stops a space
    
\IEEEcompsocitemizethanks{
\IEEEcompsocthanksitem Y. Yao and L. Zheng are with the Australian National University, Canberra, Australia. E-mail: first name.last name@anu.edu.au
\IEEEcompsocthanksitem X. Yang is with the QCraft, Santa Clara, USA. E-mail: yangxd.hust@gmail.com
\IEEEcompsocthanksitem M, Napthade is with the Capital One, San Francisco, USA, E-mail: milindinindia@yahoo.com
\IEEEcompsocthanksitem T. Gedeon is with the Curtin University, Perth, Australia, and Óbuda University, Budapest, Hungary. E-mail: tom.gedeon@curtin.edu.au
} 
%\thanks{Manuscript received April 19, 2005; revised August 26, 2015.}
}

%

% \author{Yue Yao\textsuperscript{$\dagger$},
%     Liang Zheng\textsuperscript{$\ddagger$},
%     Xiaodong Yang\textsuperscript{$\mathsection$},
%     Milind Napthade\textsuperscript{$\dagger$},
%     Tom Gedeon\textsuperscript{$\dagger$},% <-this % stops a space
    
% \IEEEcompsocitemizethanks{
% \IEEEcompsocthanksitem Y. Yao, L. Zheng and T. Gedeon are with the Australian National University. E-mail: first name.last name@anu.edu.au
% %; yue.yao@anu.edu.au; hongdong.li@anu.edu.au; liang.zheng@anu.edu.au
% \IEEEcompsocthanksitem X. Yang and M, Napthade are with the NVIDIA, Santa Clara, USA. E-mail: yangxd.hust@gmail.com, mnaphada@nvidia.com
% %\thanks{Manuscript received April 19, 2005; revised August 26, 2015.}
% }

% The paper headers
\markboth{Journal of \LaTeX\ Class Files,~Vol.~14, No.~8, August~2015}%
{Shell \MakeLowercase{\textit{et al.}}: Bare Demo of IEEEtran.cls for Computer Society Journals}
% The only time the second header will appear is for the odd numbered pages
% after the title page when using the twoside option.
% 
% *** Note that you probably will NOT want to include the author's ***
% *** name in the headers of peer review papers.                   ***
% You can use \ifCLASSOPTIONpeerreview for conditional compilation here if
% you desire.

\IEEEtitleabstractindextext{%
\begin{abstract}

%This article aims to use graphic engines to simulate a large number of training data that come with free annotations and have a possibly high resemblance to real-world data. Between synthetic and real data, there is a two-level domain gap: content level and appearance level. In contrast to the widely-studied latter, which measures the difference in image style, the former focus on content mismatch in attributes such as object placement and lighting, which lie the foundational disparity of images but are poorly learned. To address the content-level misalignment, we propose an attribute descent approach that automatically and alternatively optimizes these attributes to let synthetic data approximate real-world datasets. In verifying our method, we focus on object-centric datasets and tasks, where objects in interest take up a major part of an image. This allows us to have a relatively small attribute space, and the optimization of each attribute gives sufficiently obvious supervision signals. We collected a new synthetic asset VehicleX, and reformatted and reused existing assets ObjectX and PersonX, respectively. Extensive experiments on image classification and object re-identification tasks confirm that the simulated datasets can be effectively used in three scenarios: unsupervised domain adaptation, training data augmentation and dataset bias visualization.

This article aims to use graphic engines to simulate a large number of training data that have free annotations and possibly strongly resemble to real-world data. Between synthetic and real, a two-level domain gap exists, involving content level and appearance level. While the latter is concerned with appearance style, the former problem arises from a different mechanism, \ie content mismatch in attributes such as camera viewpoint, object placement and lighting conditions. In contrast to the widely-studied appearance-level gap, the content-level discrepancy has not been broadly studied. To address the content-level misalignment, we propose an attribute descent approach that automatically optimizes engine attributes to enable synthetic data to approximate real-world data. 
% Specifically, we model the distribution of each attribute using a Gaussian Mixture Model (GMM). We precisely manipulate GMM parameters of attributes one at a time in the graphic engine to minimize the distribution discrepancy between the synthesized and real data, where discrepancy is measured by the Fr\'{e}chet Inception Distance (FID). 
We verify our method on object-centric tasks, wherein an object takes up a major portion of an image. 
In these tasks, the search space is relatively small, and the optimization of each attribute yields sufficiently obvious supervision signals. 
We collect a new synthetic asset VehicleX, and reformat and reuse existing the synthetic assets ObjectX and PersonX. Extensive experiments on image classification and object re-identification confirm that adapted synthetic data can be effectively used in three scenarios: training with synthetic data only, training data augmentation and numerically understanding dataset content. 

\end{abstract}

% Note that keywords are not normally used for peerreview papers.
}

\maketitle

\IEEEdisplaynontitleabstractindextext
\IEEEpeerreviewmaketitle

\begin{figure}[t]
\begin{center}
	\includegraphics[width=1\linewidth]{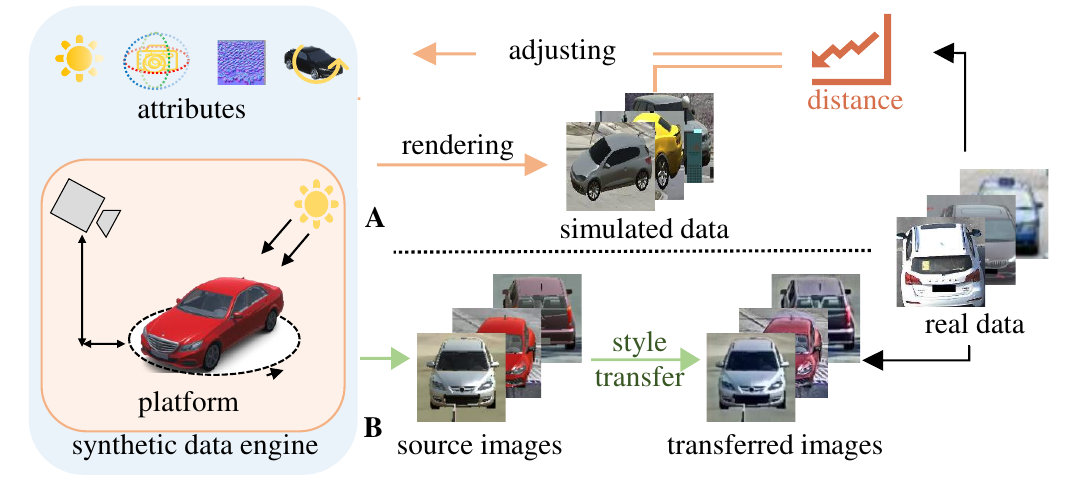}
\end{center}
\caption{
 Domain adaptation on the content level (\textbf{A}) and appearance level (\textbf{B}). With real data as a target, content-level domain adaptation optimizes image content attributes such as camera viewpoint and illumination. In comparison, appearance-level and feature-level domain adaptation typically use image style transfer (shown) or feature alignment (not shown). Specifically, given a list of attributes and their values, we use the renderer Unity for training data synthesis. Afterward, we compute the FID between the synthetic and real images to reflect their distribution difference. By updating the values of attributes using the proposed attribute descent algorithm, we can minimize the FID along the training iterations. We use the optimized attributes to generate synthetic data, which % and additionally apply style transfer. 
 %These adapted synthetic data 
 can be used to replace or scale up real-world training data.
}
\label{fig:system_flow}
\end{figure}

\IEEEraisesectionheading{\section{Introduction}\label{intro}}
\IEEEPARstart{D}{ata} synthesis that can be conveniently performed in graphic engines provides invaluable convenience and flexibility for the computer vision community~\cite{richter2016playing,sakaridis2018semantic,ruiz2019learning,tremblay2018training,sun2019dissecting}. A large number of training data can be synthesized under different combinations of environmental factors from a small number of 3D object models. Despite this convenience, a large domain gap exists between synthetic data and real-world data, which can substantially decrease accuracy when a model trained on synthetic data is tested with real data~\cite{kar2019meta,ruiz2019learning}. The domain gap problem can be addressed from the \textbf{appearance level}~\cite{kar2019meta} or the \textbf{content level} (Fig.~\ref{fig:system_flow}). The former often translates image style by using a neural network~\cite{hoffman2018cycada}, whereas the latter modifies the image content via manipulating editable attributes directly in graphic engines. 
% For statement simplicity this paper categorizes feature-level alignment methods into appearance alignment.

We are motivated by the following considerations. First, collecting large-scale real-world datasets with manual labels is expensive. For example, in a multi-camera system like object re-identification (re-ID), an object must be associated across multiple different cameras to obtain ground truth labels: this process is extremely difficult and laborious because objects usually vary in appearance under different cameras. Privacy and data security concerns further add overheads to this process.  

Second, the domain gap between datasets exists not only on the appearance level~\cite{hoffman2018cycada}, but also on the content level~\cite{kar2019meta}. Take vehicle re-ID as an example. Images in the VehicleID dataset~\cite{liu2016deep} mainly exhibit car rears and fronts, whereas the VeRi-776 dataset~\cite{liu2016large} covers more diverse viewpoints. This difference in viewpoints is considered an example of content-level discrepancy between datasets. As a result, models trained on VehicleID~\cite{liu2016deep} have a significant accuracy drop when tested on VeRi-776~\cite{liu2016large}. Existing domain adaptation methods on the appearance level can alleviate this problem and improve accuracy, but are essentially incapable of handling the content differences. %As to be shown in our experiment, content-level alignment is orthogonal and complementary to existing domain adaptation techniques on the appearance level. 

%--------------------------------------------figure1----------------------------------------------

Given these considerations, this article proposes an attribute descent algorithm to generate synthetic training data 1) in a large scale and inexpensive manner, 2) with a reduced content domain gap with real-world data. In a nutshell, as shown in Fig.~\ref{fig:system_flow} \textbf{A}, the proposed attribute descent algorithm can automatically configure simulator attributes, such as camera viewpoint, lighting direction, and object placement, so that the synthesized data are close to real-world data as measured by the Fr\'{e}chet Inception Distance (FID)~\cite{heusel2017gans}. These optimized attributes are subsequently used for training set generation or augmentation, where style-level image translation can be leveraged to further reduce domain discrepancy. To make the above process happen, we define a controllable simulation environment in the graphic engine with editable object and environment attributes, allowing us to generate large training sets by varying these attributes. 

We evaluate the effectiveness of synthetic data generated by attribute descent on \emph{object-centric} tasks: image classification, person re-ID and vehicle re-ID. In these tasks, an image usually contains a single object of interest which is placed approximately in the image center (through either automatic detection or manual cropping). Specifically, image classification aims to distinguish different classes for an input object of interest, while object re-ID targets at differentiating object (person, vehicle) identities. For each of the tasks, we build a synthetic \emph{object-centric} filming scenario, which contains an object of interest, a camera and a direct lighting source. Compared with complex-scene applications such as semantic segmentation, object-centric tasks require fewer 3D object assets and have simpler layout relationships between assets, smaller environments and fewer attributes to be controlled. These favorable properties simplify the problem addressed and enable in-depth method analysis. More discussions of the application scope of our method are provided in Section~\ref{sec:method_disscussion}.   

We identify three scenarios in which data synthesized by attribute descent can be effectively applied: training with adapted synthetic data only, augmenting real training data with synthetic data, and visualizing content-level dataset bias. \textbf{First}, after optimizing synthetic data to approximate target distribution (real world), we directly use synthetic data to train object-centric task models for recognition or re-ID. \textbf{Second}, we add the optimized synthetic data to real-world data to enlarge the training set, which is used for training the task model and improving accuracy. In both applications, we demonstrate that optimized synthetic data are superior to those randomly generated and that the inclusion of the former consistently enhances real-world training data. Furthermore, the two applications suggest the existence of content bias in object-centric datasets, and a \textbf{third} and interesting application is understanding dataset content by using obtained attribute value distributions. For example, we visualize that vehicle orientations on a normal two-way road are usually bi-modal, exhibiting two major angles, whereas those in an intersection are more diverse.  

Experimentally we show that attribute descent is superior to using random attributes for training set synthesis, when the training is either with synthetic data only or has both synthetic and real data. Moreover, we compare attribute descent with existing gradient-free optimization techniques, namely Bayesian optimization, reinforcement learning, evolutionary algorithm, and random search. We observe that attribute descent leads to a consistently lower domain gap between the generated training set and target set, consequently yielding the model higher accuracy than the competing methods. Additionally, we report that attribute descent promotes more stable convergence than the competing methods. The above indicates the effectiveness of attribute descent for syn2real content-level domain adaptation. 

%We conduct a comparative analysis of attribute descent with established gradient-free optimization techniques, namely Bayesian optimization, reinforcement learning, evolutionary algorithms, and random search. Through our experiments, we demonstrate the distinct advantages of attribute descent in accurately approximating target distributions, consequently enabling the generation of a superior target-specific synthetic training dataset. Additionally, our empirical findings indicate that attribute descent promotes stable convergence of the utilized loss function. Given these compelling attributes, we propose that attribute descent presents itself as a straightforward yet effective baseline for syn2real content-level domain adaptation.  }

This article is an extension of our conference publication~\cite{yao2019simulating}. Three major differences are presented. First, while only vehicle re-ID is studied in \cite{yao2019simulating}, our current system is capable of additionally improving person re-ID and image classification tasks. For these new tasks, we collect/extend new/existing 3D assets in the person and generic object domains and customize them into our pipeline. 
Second, comprehensive geometric modeling of object-camera placement. This introduces an additional in-plane rotation attribute that allows us to model object orientation more flexibly, particularly for generic objects. Third, by approximating real-world datasets with synthetic counterparts, we show that certain aspects of dataset content in real-world data can be numerically understood and visualized by this synthetic proxy. Interesting experimental observations are presented and discussed. 
 
%--------------------------section2--related work ------------------------------------
\section{Related Work}
\label{sec:related work}

\textbf{Realistic appearance generation and domain adaptation.} Domain adaptation is often achieved by reducing the domain gaps between distributions. To date, a majority of works in this field has focused on discrepancies in \textbf{appearance} or \textbf{feature level}. For the former, a considerable body of research use the cycle generative adversarial network (CycleGAN)~\cite{zhu2017unpaired} and its variants to reduce the appearance gap~\cite{hoffman2018cycada,shrivastava2017learning,deng2018image,li2019bidirectional,chen2019crdoco}. The latter models the dependence between two domains by means of feature-level statistics~\cite{kouw2016feature}, and many moment matching schemes are studied to learn a shared feature representation~\cite{sun2016return,long2015learning,tzeng2014deep,peng2019moment,zhang2018aligning,zou2020dgnetpp}. Although these works are shown to be effective in reducing the appearance or feature domain gap, a fundamental problem remains to be solved, \emph{i.e.}, the content difference. 

%---------------------table1--synthetic-dataset------------------------------------------------

\begin{table}
\centering
\caption{Statistics of synthetic datasets used in this article and their comparison with several existing synthetic datasets. Note that ObjectX and VehicleX are newly introduced. ``Attr'' denotes whether a dataset has attribute labels (\emph{e.g.,} orientation). The number of synthetic images here is not compared, because a potentially unlimited number of images can be created by these engines. Of note, for object re-ID, the number of models is equal to the number of IDs.}
\label{table:Datasets}
\setlength{\tabcolsep}{1.2mm}{
\begin{tabular}{c|l|c|c|c} 
\Xhline{1.2pt}
Task                                               & Dataset       & \multicolumn{1}{l|}{\#Models } & \#Classes (IDs)             & Attr.~  \\ 
\hline
\multirow{2}{*}{Image cla.~~}                      & VisDA Source~\cite{peng2017visda} & 1,907                                & 12               &   \xmark         \\
                                                    & ObjectX        & 1,400                                & 7 &    \cmark        \\ 
\hline
\multirow{3}{*}{Person re-ID}                       & RandPerson~\cite{wang2020surpassing}     & 8,000                                & 8,000               &  \xmark          \\
                                                    & UnrealPerson~\cite{zhang2021unrealperson} & 3,000                                & 3,000              &   \xmark         \\
                                                    & PersonX~\cite{sun2019dissecting}        & 1,266                                & 1,266  &     \cmark       \\ 
\hline
\multicolumn{1}{l|}{\multirow{2}{*}{Vehicle re-ID}} & PAMTRI~\cite{tang2019pamtri}         & 402                                  & 402                &    \cmark        \\
\multicolumn{1}{l|}{}                               & VehicleX       & 1,362                                & 1,362 &     \cmark       \\
\Xhline{1.2pt}
\end{tabular}}
\end{table}

\textbf{Learning from simulated 3D data.} Data simulation is an inexpensive way of increasing a training set scale while providing accurate image labels, flexibility in content generation and high resolution. Learning from simulated data finds it applications in image recognition~\cite{peng2017visda}, re-identification~\cite{sun2019dissecting,tang2019pamtri}, semantic segmentation~\cite{hoffman2018cycada,gaidon2016virtual,xue2021learning}, navigation~\cite{kolve2017ai2} and detection~\cite{kar2019meta,hou2020multiview}. Existing knowledge or estimation of data distribution is usually required during data synthesis in the graphic engine. Some applications directly take what is presented in existing video games such as GTA5, which have pre-defined scenes and objects~\cite{richter2016playing,xiang2020unsupervised,hu2019sail,wang2019learning,peng2017visda,doan2018g2d}. Others manually create their own simulation environments and objects~\cite{dosovitskiy2017carla,gaidon2016virtual,kolve2017ai2,deitke2020robothor,mueller2017sim4cv}, and find it is beneficial to use random attribute within a reasonable range to create random content \cite{tremblay2018training,tobin2017domain,mozifian2020intervention}. However, despite being called ``random'', the range of random variables still must be specified manually according to experience. Instead, we aim to learn attribute distributions more automatically or with less human experience.

\textbf{Automatic 3D content creation.} Many works aim to automatically create \emph{realistic} 3D models~\cite{jones2020shapeassembly,yin20213dstylenet}, focusing on intrinsic properties of 3D models such as backbones, geometry and surface. The studied object models include faces~\cite{shi2019face}, persons~\cite{zhang2020generating,paschalidou2021neural}, vehicles~\cite{yin20213dstylenet} and furniture~\cite{jones2020shapeassembly}. Departing from these works that mainly study object synthesis, we aim to use the optimized 3D models for downstream task training. %  or 3D environments~\cite{kar2019meta}, focusing on intrinsic properties of 3D models such as backbones, geometry and surface.  For example, Jones \emph{et a.} The former study models such as faces~\cite{shi2019face}, persons~\cite{zhang2020generating,paschalidou2021neural}, vehicles~\cite{yin20213dstylenet} and furniture~\cite{jones2020shapeassembly}. The latter has applications like driving simulation~\cite{chen2021geosim}, street scene generation~\cite{kar2019meta} and indoor scene generation~\cite{paschalidou2021atiss}. \red{3For example, aiming to generate arbitrary face, Shi \etal define a bone-driven 3D face model \cite{shi2019face}; 
%Targeting to generate various street scene, Chen \etal and Kar \etal define a street layout model with probabilistic scene grammar from attributes \cite{chen2021geosim,kar2019meta}. 
%we, aspiring to deal with diverse object-camera placement in comparison, delicately define a camera model and editable attributes for object-centric tasks as to be discussed in Section \ref{sec:Camera_Model}. }

\textbf{Content creation for task model training.} Several recent studies have attempted to automatically generate 3D content \emph{for training task models}~\cite{kar2019meta,devaranjan2020meta,xue2021learning,ruiz2019learning}. They are closest to our work. For example, Kar \etal and Devaranjan \etal simulate traffic for training vehicle detection networks~\cite{kar2019meta,devaranjan2020meta}. Xue \etal and Ruiz \etal construct street scenes for semantic segmentation model training~\cite{xue2021learning,ruiz2019learning}. Many of these works use reinforcement learning (RL) based algorithms to optimize attributes \cite{kar2019meta,devaranjan2020meta,xue2021learning,ruiz2019learning}. In our article, we find RL less effective in object-centric tasks, in which a relatively small number of attributes needed to be optimized. In this sense, our method is an alternative to existing ones and particularly effective in object-centric tasks (see comparisons in Table~\ref{tab:Resultvehiclex2vehicleID_unsupervised}). 
 
In \textbf{object-centric vision tasks}, a single object of interest usually appears in an image, because manual cropping or detection is used to place the object approximately in the image center. Specifically, the basic image classification problem aims to distinguish different classes for an input image~\cite{deng2009imagenet,lin2014microsoft,he2016deep,simonyan2014very}. Another task that we consider is object re-ID~\cite{zheng2015scalable,ristani2016MTMC,liu2016large}, which has many robust systems proposed recently~\cite{khorramshahi2019dual,wang2017orientation,tang2019pamtri,zhou2018aware}. When experimenting on these object-centric tasks, we adopt existing architectures and loss functions with no bells and whistles. 

\textbf{Dataset bias} is a critical reason for compromised model performance \cite{torralba2011unbiased}. For example, many classification datasets~\cite{deng2009imagenet,lin2014microsoft,kuznetsova2020open} are collected from public user repositories, such as Flickr, which may have content bias in terms of object placement, background, rotation, occlusion, lighting conditions \emph{etc}. The bias may explain why models exhibit lower accuracy on test sets that have a different content bias from the training set \cite{barbu2019objectnet}. There are some existing approaches for dataset bias visualization. For example, t-distributed stochastic neighbor embedding (t-SNE) can exhibit \emph{feature-level} data distributions~\cite{van2008visualizing}. In comparison, we aim to visualize bias through attributes, % a more interpretable way, 
\ie obtaining attribute distributions numerically and drawing them in graphs. %visualizing the distribution of attributes. 

\begin{figure}[t] 
    \centering
    \begin{center}
        \includegraphics[width=1\linewidth]{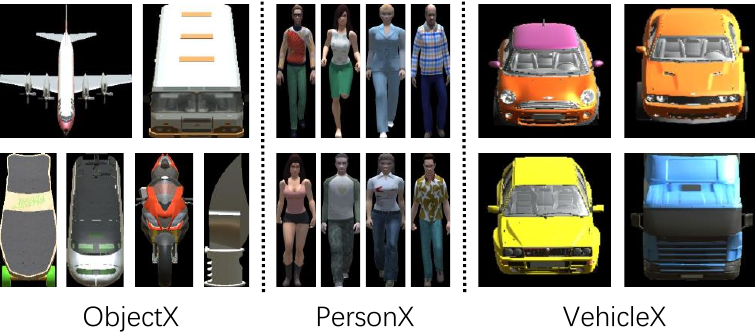}
        \caption{Sample 3D models for ObjectX, PersonX~\cite{sun2019dissecting}, and VehicleX~\cite{yao2019simulating}, which are used for synthesizing images for image classification, person re-ID, and vehicle re-ID, respectively.}
        \label{fig:3D_models}
    \end{center}
    % \vspace{-3.1em}
\end{figure}

Similarly, depending on the camera condition, location and environment, existing object re-ID datasets usually have their own distinct characteristics or bias~\cite{sun2019dissecting}. For example, vehicle images in the VehicleID dataset~\cite{liu2016deep} are either captured from car front or back, whereas the VeRi dataset~\cite{liu2016large} includes a much wider range of viewpoints. Beyond dataset-dataset differences, large differences also exist between cameras in a single dataset~\cite{zhong2018camstyle}. For example, a camera filming a cross road naturally has more vehicles orientation than a camera on a straight road. In this article, we leverage such characteristics to learn attributes for each camera and simultaneously visualize the attribute distributions. 
%--------------------------section3--related work ------------------------------------

\section{Simulation Environment}

Overall, we aim to build a simulation environment with editable attributes for building large-scale synthetic datasets that are close to the real world. To achieve this, we collect a large number of 3D objects (Section~\ref{sec:3D_Asset-Acquisition}), build a camera model in the graphic engine, define a set of editable attributes (Section~\ref{sec:Camera_Model}), and acquire synthetic images by varying these attributes (Section~\ref{sec:filming_process}).

\subsection{3D Asset Acquisition}
\label{sec:3D_Asset-Acquisition}

For object-centric tasks, we use 3D assets from three sources. Their details are provided below as well as in Table~\ref{table:Datasets}, and sample assets are shown in Fig.~\ref{fig:3D_models}. 

We reformat \textbf{ObjectX} from ShapeNet-V2~\cite{chang2015shapenet} to simulate classification data. Similarly to ImageNet~\cite{deng2009imagenet}, ShapeNet organizes 3D shapes according to the WordNet hierarchy~\cite{miller1995wordnet}. From ShapeNet, we select the classes that are also included in the VisDA target dataset (containing real-world images)~\cite{peng2017visda}, which are used as our target data. This amounts to 7 classes and 200 models which are randomly selected for each class. During pre-possessing, models in each class are aligned in the same direction and scaled to a uniform size. Table~\ref{table:Datasets} shows the statistics of ObjectX, and Fig.~\ref{fig:3D_models} visualizes some 3D shapes we collected for categories \emph{airplane}, \emph{bus}, \emph{skateboard}, \emph{train}, \emph{motorcycle}, and \emph{knife}.

We use \textbf{PersonX} to simulate person re-ID data. This asset is introduced by Sun \etal~\cite{sun2019dissecting} and has 1,266 different manually constructed person models (identities), including 547 females and 719 males. PersonX models are hand-crafted by professional 3D modelers, with a special focus on appearance diversity. To explain, the backbone models of PersonX have various ages, hairstyles, and skin colors. For each backbone model, the clothes are also chosen from a diverse range including T-shirts, skirts, jeans, shorts, pants, slacks, \etc. With real-world-like textures, these clothes have good visual authenticity. Furthermore, a person can take various actions (\eg walking, running) when being filmed. Viewpoint alignment and scaling are performed for 3D objects in PersonX. 

\textbf{VehicleX}, introduced in our conference paper~\cite{yao2019simulating}, is used to simulate data for the vehicle re-ID task. It has a wide range of simulated backbone models and textures and adapts well to the variance of real-world datasets. Specifically, it has 272 backbones which are hand-crafted by professional 3D modelers. The backbones include 11 mainstream vehicle types including sedan, SUV, van, hatchback, MPV, pickup, bus, truck, estate, sportscar and RV. On the basis of these backbones, we obtain 1,362 vehicle identities by adding various colored textures or accessories. A comparison of VehicleX with an existing vehicle re-ID dataset (\ie PAMIRI~\cite{tang2019pamtri}) is presented in Table~\ref{table:Datasets}. VehicleX is three times larger than the synthetic dataset PAMTRI in the number of identities, and, similar to ObjectX and PersonX, can potentially render an unlimited number of images by varying the attributes. In VehicleX, similarly to PersonX, vehicle models are aligned to their frontal views and properly scaled. Note that for object (person and vehicle) re-ID, the testing procedure is retrieval; therefore, synthetic data need not to have the same classes with real-world target data. VehicleX has been used as training data in the AI city challenge in CVPR 2020 and CVPR 2021\footnote{https://www.aicitychallenge.org/}.

\begin{figure}[t] 
    \centering
    \begin{center}
        \includegraphics[width=1\linewidth]{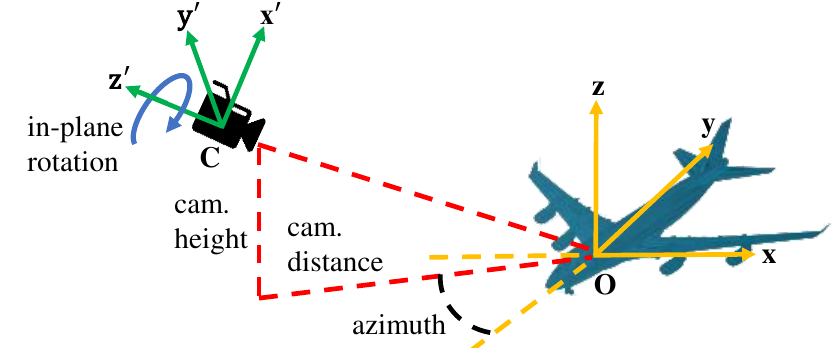}
        \caption{The camera model in our simulation environment. Given fixed focal length and resolution, the transformation matrix between the world coordinate system \textit{O} and camera coordinate system \textit{C} is determined by the azimuth, camera height, camera distance, and in-plane rotation. }
        \label{fig:Cam_model}
    \end{center}
    % \vspace{-1em}
\end{figure}

\subsection{Camera Model}
\label{sec:Camera_Model}

To capture 2D images of the 3D objects, we create a camera model, as illustrated in Fig.~\ref{fig:Cam_model}. Let the world coordinate system be \textit{O}$(x, y, z)$, in which the origin is the center of the 3D object, and the $y$ axis points toward a certain direction of each class of objects (\emph{e.g.,} the frontal view of airplanes). The camera coordinate system \textit{C} is denoted by $(x', y', z')$. To ensure that the target is filmed and to simplify the camera model, we hold a prior that the camera is always facing the 3D object, \emph{i.e.,} towards the center of the world coordinate system (or object center). Thus, we define the $z'$ axis as pointing away from the center of the object. 

\begin{figure*}[t] 
    \centering
    \begin{center}
        \includegraphics[width=1\linewidth]{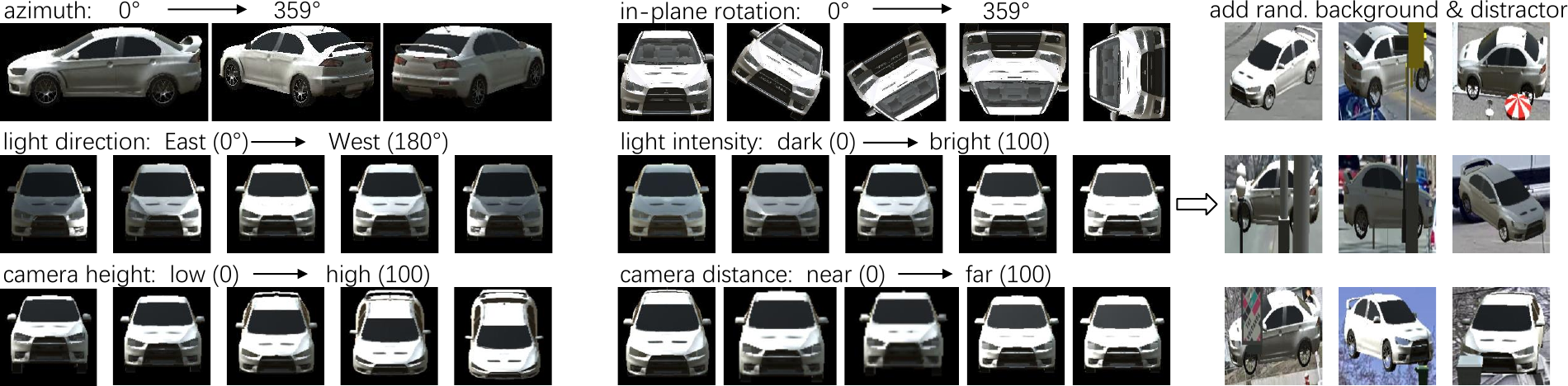}
        \caption{(\textbf{Left} and \textbf{Middle:}) Illustration of editable attributes. We can rotate the object (azimuth) and camera (in-plane rotation), edit light direction and intensity, and change the camera height and distance. Value and range are shown for each attribute. % for each attribute  the bracket correspond to the attribute values in Unity.
        (\textbf{Right:}) After attribute optimization, we further add random backgrounds and occlusions to images when synthesizing the final training dataset. %to the attribute-adjusted objects when they are used in the re-ID model. 
        }
        \label{fig: editing}
    \end{center}
    \vspace{-1.0em}
\end{figure*}

We proceed to consider extrinsic and intrinsic parameters that transform the world coordinate system into the camera coordinate system. \textbf{Extrinsic parameters} are determined by the rotation matrix $R$ and translation vector $T$. $R$ includes object rotation and camera rotation. Object rotation includes the azimuth $\alpha$ and elevation, which can be represented by the function of camera height $h$ and camera distance $d$. Camera rotation involves only in-plane rotation, because we assume that our camera always faces the object. Thus, the rotation matrix $R$ is a function of $\alpha$, $h$, $d$ and $\theta$, and can be written as $R(\alpha, h, d, \theta)$. The translation vector $T$ is determined directly by camera height $h$ and camera distance $d$, and thus can be written as $T(h, d)$. For \textbf{intrinsic parameters}, we use a fixed focus length $f$ and image resolution $\gamma$. For example, we set the world coordinate system to a resolution of $1,920 \times 1,080$ pixels when capturing vehicle images. Given the above extrinsic parameters and intrinsic parameters, the projection matrix can be written as,
\begin{equation}
\begin{split}
P =\underbrace { \begin{bmatrix}
\gamma f & 0 & 0 \\ 
 0 & \gamma f  & 0 \\ 
0 & 0 & 1 
\end{bmatrix}}_{\rm intrinsic \,\, parameters}\underbrace{ \begin{bmatrix}
R(\alpha, h, d, \theta)_{3 \times 3} & T(h, d)_{3 \times 1} \\ 
0_{1 \times 3} &  1
\end{bmatrix} }_{\rm extrinsic \,\, parameters}.
\end{split}
\label{eq:cam_model}
\end{equation}
Thus, given focal length $f$ and resolution $\gamma$, the camera model has 4 remaining attributes to be configured: azimuth $\alpha$, camera height $h$, camera distance $d$ and in-plane rotation $\theta$. These left 4 attributes determine the diverse object-camera placement, thus allowing us to film a variety of images. This camera model is an improved version from our conference paper \cite{yao2019simulating}. The additionally introduced attribute ``in-plane rotation'' allows us to more flexibly model objects that do not usually ``stand'' on the ground plane. 
 %Note that compared with our conference version~\cite{yao2019simulating}, in-plane rotation $\theta$ is newly added, in a purpose of having more diverse object orientations. 
% In other word, 

\subsection{Configurable Attributes}
\label{sec:Configurable_attr}
Our system has a total of 6 editable attributes, which are considered to be influential on the training set quality and the subsequent testing accuracy. They include azimuth, camera height, camera distance and in-plane rotation, as mentioned in Section~\ref{sec:Camera_Model}, plus two lighting attributes: light direction and light intensity. Several details regarding these attributes are provided below. 
\begin{itemize}[noitemsep,topsep=5pt]
    % \item \red{\underline{Object orientation}}
	\item \underline{Azimuth} represents the horizontal viewpoint of an object and takes a value between $0^{\circ}$ and $359^{\circ}$. 
    \item \underline{In-plane rotation} controls camera rotation on the $z'$ axis. Its value is also between $0^{\circ}$ and $359^{\circ}$. 
	\item \underline{Camera height} describes the vertical distance of the camera from the ground, from near (numerical value $0$) to far (numerical value $100$). 
	\item \underline{Camera distance} determines the horizontal distance between the camera and the object center. This factor, taking values from low (0) to high (100), strongly affects on the imaging resolution (the resolution of the entire camera view is set to 1920$\times$1080). Imaging viewpoint is impacted by the joint effects of camera height and distance.  
	\item \underline{Light direction} is part of the environment settings. We assume directional parallel light, and the light direction is modeled from east ($0^{\circ}$) to west ($180^{\circ}$). 
	\item \underline{Light intensity} is another a critical attribute influencing task performance by creating reflections and shadows. We manually define a reasonable range for intensity, from dark ($0$) to light ($100$). 
\end{itemize}

Note that the pre-defined ranges are broad and involves little human assumption. For example, light intensity ranges from fully light to dark. From the ranges, we aim to find their optimal values for effective training dataset construction. Moreover, these 6 attributes (examples shown in Fig.~\ref{fig: editing}) are found important for the three object-centric tasks. For other tasks such as semantic segmentation and object detection, which involve complex environments and multiple objects, other influential attributes, such as object (relative) locations, are influential. In this article, we try to keep our investigation focused and in depth and leave the more complex applications for future work.

\subsection{Image Capturing Process}
\label{sec:filming_process}

In this section, we describe how to film the 3D objects after varying these editable attributes. To establish controllability, we build a \textbf{Unity-Python interface} by using the Unity ML-Agents toolkit~\cite{juliani2018unity}, which creates a data transmission path between Unity and Python. Specifically, given a list of parameters from Python, Unity accordingly sets up the camera and environment, then takes the picture and obtains the bounding box precisely according to the object size. The bounding box is then sent back to Python for further processing. Our API allows users to easily obtain rendered images without expert knowledge of Unity. The source code of this API has been released\footnote{https://github.com/yorkeyao/VehicleX}.
  
We also consider background and occlusion setups in our system. Specifically, during the attribute distribution optimization process with attribute descent, we render images with black backgrounds and no occlusions. After the optimization is complete, we generate synthetic training data using the optimized attributes while adding random background images and occlusions. For example, when synthesizing vehicle re-ID training data, we add background images from the CityFlow dataset~\cite{tang2019cityflow}, and occluders such as lamp posts, billboards and trash cans. Our preliminary experiments show this strategy increases the diversity of the synthesized training data, consequently enabling robust training of the task model. For example, when using synthetic data only and targeting vehicleID, training on synthetic images with black backgrounds and attribute descent produces 24.12\% mAP. The further incorporation of random backgrounds and occlusions yields a notable increase of +11.21\% mAP over the training dataset that uses black backgrounds. This strategy allows us to train models that are more robust to various backgrounds and occlusions.

\section{Proposed Method}\label{sec:attribute_descent}

% \subsection{Objective}

% \textcolor{blue}{Overall, we aim to optimize the synthetic training data so that it can be visually similar to our target and thus can train a model that has high accuracy. To achieve this, we }

\subsection{Attribute Distribution Modeling}\label{sec:att_model}

We model the distribution of the 6 attributes (Section~\ref{sec:Configurable_attr}) using Gaussian Mixture Models (GMM). The primary rationale is that for real-world datasets, these attributes usually follow specific patterns. In the \emph{image recognition task}, patterns often exist with regard to specific categories. For example, photographs of the category \emph{clock} are usually taken from the front, not the side or back, where the Gaussian distribution would be helpful. Likewise, in the \emph{re-ID task}, the camera position is usually fixed, so that the camera height and distance of images captured by a certain camera are generally uni-modal. Moreover, pedestrians and vehicles usually move along predefined trajectories, \eg footpaths for pedestrians and traffic lanes for vehicles. Because multiple moving directions exist on a single path, the object orientations (\ie the azimuth) of these re-ID data exhibit multi-modal distributions. Of note, this modeling strategy has also been adopted by Ruiz \emph{et al.} \cite{ruiz2018learning}. In addition, using Gaussians and their mixtures to model the distribution of attributes allows us to conveniently analyze and visualize datasets. %We show attributes that minimize the distribution difference between datasets can produce effective synthetic training data. 

\begin{figure*}[t] 
    \centering
    \begin{center}  
        \includegraphics[width=1\linewidth]{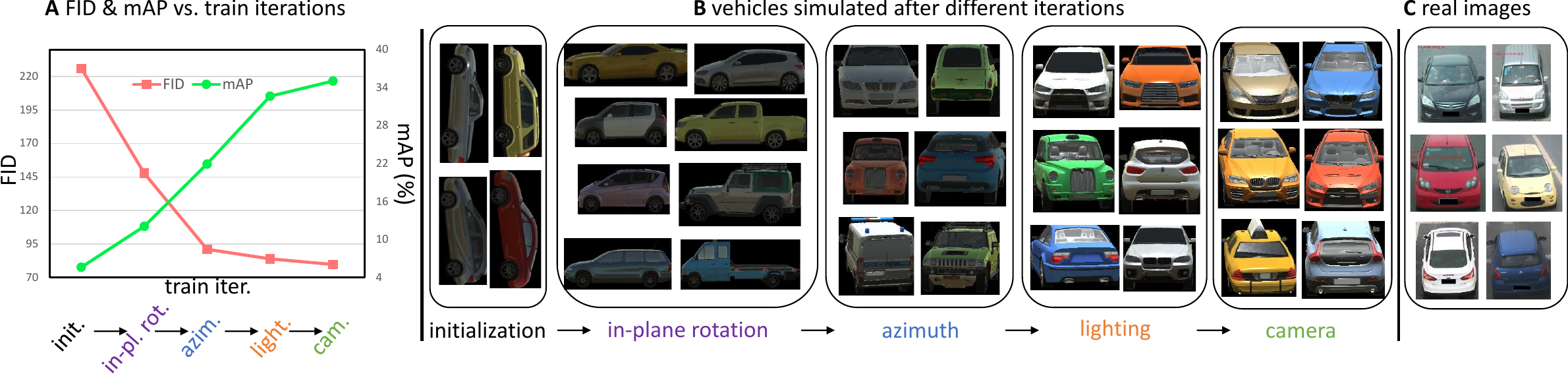}
        \caption{Visualization of the attribute descent process on the VehicleID~\cite{liu2016deep} dataset. (\textbf{A:}) FID and task accuracy mean average precision (mAP) (\%) \emph{vs.} training iterations. We observe that the FID between synthetic data and target real data (\textbf{C}) successively decreases and that mAP on the target domain gradually increases (higher is better). ``In-pl. rot.'' denotes in-plane rotation, ``lighting'' represents light direction and intensity, and ``cam.'' means camera height and distance. (\textbf{B:}) we show the synthetic vehicles after optimizing each attribute in a certain epoch. We initialize attributes by setting the in-plane rotation to $90\degree$, orientation to the left (0), light intensity to dark (0), light direction to east (0), camera height to the same level as the vehicle center (0), and camera distance to medium (50). Along the iterations, the content of these synthetic images becomes increasingly similar to (\textbf{C}) the target real-world images.
        }
        \label{fig:training}
    \end{center}
    % \vspace{-1em}
\end{figure*}

Although a GMM is parameterized by its mean and covariance, our preliminary shows that the change in covariance produces a less prominent supervision signal, \emph{i.e.,} has a weaker effect on task accuracy than the mean. This finding is understandable because the mean parameters encode how the majority of images appear and more directly reflect the dataset distribution gap. Although our method can optimize covariance, doing so would significantly increase the search space without notable accuracy gain. Based on the above considerations, we simplify the covariance matrix into a diagonal matrix whose diagonal elements are pre-defined. 

Formally, let $\bm{A}= [A_1, A_2,..., A_K]^T\in \mathbb{R}^K$ represent attributes discussed in Section \ref{sec:Configurable_attr}, where $K$ stands for the total number of attributes. $A_i\in \mathbb{R}, i = 1,...,K$ is the random variable representing the $i$th attribute. As discussed in Section~\ref{sec:Configurable_attr}, we have $K=6$, where $A_1$, $A_2$, $A_3$, $A_4$, $A_5$, and $A_6$ denote azimuth, in-plane rotation, camera height, camera distance, light direction, and light intensity, respectively.

Let $\bm{\theta}= [\theta_1, \theta_2,..., \theta_M]^T$ represent all the learnable parameters, where $M$ is the number of these learnable parameters, and $\theta_i\in \mathbb{R}, i = 1,...,M$, is a single learnable parameter (\ie mean, as variance is fixed) of one mixture component. Thus, the $\bm{\theta}$ can be viewed as a concatenated vector of parameters of all GMMs. Subsequently, we use $G(\bm{\theta})$ to denote the overall distribution parameterized by $\bm{\theta}$, which consists of $K$ GMMs. For example, we define the distribution of in-plane rotation attribute $A_1$ using a GMM consisting of three components. For each component, it has a mean value as a learnable parameter. Formally, they are denoted as $\theta_1$, $\theta_2$, and $\theta_3$. For attributes that empirically have simpler distributions, we use GMM with fewer components. For example, we define the camera distance attribute $A_2$ to follow a single-component Gaussian, which has only one learnable parameter $\theta_4$.

Furthermore, to make things easier for optimization, we let $\bm{S} = [\bm{S_1}, \bm{S_2},..., \bm{S_M}]^T$ represent a collection of search spaces for learnable parameters $\bm{\theta}=[\theta_1, \theta_2,..., \theta_M]^T$, where $\bm{S_i}\in \mathbb{R}^{d_i}, i=1,...,M$, is the search space of parameter $\theta_i$. The dimension $d_i$ of $\bm{S_i}$ is dependant on the attribute  $\theta_i$. For example, we have $\theta_1$, $\theta_2$, and $\theta_3$ are learnable parameters of attribute $A_1$ (in-plane rotation), and their search spaces $\bm{S_1} = \bm{S_2} = \bm{S_3} = [30l \mid l = 0, 1, \ldots, 11 ]^T$, which is a 12-dim vector including angles between $0^{\circ}$ and $330^{\circ}$ with an increment of $30^{\circ}$. A complete attribute distribution and search space description of all the learnable parameters is provided in Section~\ref{sec:Experimental_Details}. %corresponding to the first parameter $\theta_1$ in attribute $A_1$.}

\subsection{Optimization}

To describe the attribute descent optimization process, we introduce the objective function, detail the proposed attribute descent algorithm and then analyze its convergence property. 

\textbf{Objective function.} Given real-world target data, we aim to synthesize a dataset that has a minimal (content) distribution difference with it. Formally, we denote the sets of synthetic images and real images as $D^s$ and $D^r$, respectively, where the synthetic dataset is simulated following the distribution of vector $\bm{A}$. We further write $D^s$ as $\{\mathcal{R}({\bm{a}}_n)\}_{n=1}^{N}$, where $N$ is the number of images in $D^s$. $\mathcal{R(\cdot)}$ is the underlying rendering function of Unity using attribute vector $\bm{a}_n$ as input and producing a synthetic image $\mathcal{R}({\bm{a}}_n)$. Each attribute vector $\bm{a}_n$ is sampled from distribution parameterized by $\bm \theta$, \ie $\bm{a}_n \sim G(\bm{\theta})$. We aim to optimize $\bm{\theta}$ to minimize the distribution difference measure Fr\'{e}chet Inception Distance (FID)~\cite{heusel2017gans} between $D^s$ and $D^r$. Correspondingly, our objective function is written as:
\begin{equation}
\begin{split}
% \min_{\bm{\theta}}L_{\bm{A}\sim G(\bm {\theta})}(D_{\bm{A}}^s, D^r) = 
\bm{\theta}^{*} =  \mathop{\arg\min}_{\bm{\theta}}\mbox{FID}(D^s, D^r), 
\end{split}
\label{eq:loss}
\end{equation}
where
\begin{equation}
\begin{split}
D^s = \{\mathcal{R}({\bm{a}}_n)\}_{n=1}^{N}, \bm{a}_n \sim G(\bm{\theta}),
%\mbox{and}
\end{split}
\label{eq:data}
\end{equation}
and
\begin{equation}
\begin{split}
\mbox{FID}(D^s, D^r) = \left \| \bm{\mu}^s - \bm{\mu}^r  \right \|^{2}_{2} + \qquad  \qquad \\ \qquad \qquad \qquad
         Tr(\bm{\Sigma}^s + \bm{\Sigma}^r -2 (\bm{\Sigma}^s \bm{\Sigma}^r)^{\frac{1}{2}}).
\end{split}
\label{eq:fid}
\end{equation}
In Eq. \ref{eq:fid}, $\bm{\mu}^s \in \mathbb{R}^d$ and $\bm{\Sigma}^s \in \mathbb{R}^{d\times d}$ denote the mean and covariance matrix, respectively, of the image descriptors of the synthetic dataset $D^s$, and $\bm{\mu}^r$ and $\bm{\Sigma}^r$ are those of the real-world dataset $D^r$, respectively. $d$ is the dimension of the image descriptors, which are extracted by the InceptionV3 model~\cite{szegedy2016rethinking} pre-trained on the ImageNet dataset~\cite{deng2009imagenet}.

The benefit of using FID as the loss function is two-fold. First, it does not require ground truths of the target real-world dataset $D^r$ and instead can be computed using target images only. This obviously saves human labeling efforts. Second, because the FID loss is computed between two sets of images, it is much faster to compute than existing methods using the task loss \cite{kim2017learning,kar2019meta}. The latter requires training the task model (\eg object detection~\cite{girshick2015fast}) with the generated training set and uses the obtained model task performance as loss, which is time-consuming.

% \textcolor{blue}{The usage of FID loss has multiple benefits. First, its calculation 
%  requires images only. Thus we are not required to know the ground truth labels of the real dataset, which usually takes human efforts to manually label them. Second, compared with existing methods~\cite{kim2017learning,kar2019meta} that use task loss to guide the content-level domain adaptation, FID loss enables faster calculation, as it does not require performing actual model training, a well-known time-consuming process. 
% }

It is important to note that the objective function (Eq. \ref{eq:loss}, Eq. \ref{eq:data} and Eq. \ref{eq:fid}) is non-differentiable with respect to $\bm\theta$, primarily because the rendering function (through the 3D engine Unity) is not differentiable. As such, we cannot perform optimization by directly using gradient descent.

\textbf{Attribute descent algorithm.} We are motivated by coordinate descent, a classical optimization method that can work under derivative-free scenarios~\cite{wright2015coordinate}. To find a local minimum, it selects a coordinate direction to perform a step-by-step search and iterates among coordinate directions. Compared with grid search that considers the entire search space, coordinate descent has significantly reduced search space and thus running time.
% based on the hypothesis that parameters are relatively independent. 
\begin{algorithm}[t]  
  \caption{Attribute Descent}  
  \label{alg:Framwork}  
  \begin{algorithmic}[1]
    \State \textbf{Input:} Initialized learnable parameters $\bm{\theta}=\left[\theta_1^0,\cdots,\theta_M^0\right]^T$, search space $\bm{S}=[\bm{S_1}, \bm{S_2},..., \bm{S_M}]^T$, %for each element in $\bm{\theta}$, 
    rendering function $\mathcal{R}(\cdot)$ and target dataset $D^r$.
    \State \textbf{Hyperparameteters:} $J$ epochs for attribute descent and synthetic dataset size $N$. 
    \State \textbf{Begin:}
    \State Optimal = $\infty$; \Comment{Initialize the optimal FID value}
    \For{ $j = 1$ \textbf{to} $J$ } \Comment{Number of epoch J}
        % \State $\bm{\mu^{k}} = \bm{\mu^{k-1}}$
        % \State Select a random set of attributes 
        \For{$i = 1$ \textbf{to} M} \Comment{Enumerate parameters}
            \For{$z \in \bm S_{i}$} \Comment{Traverse search space}
                \State $D^s = \{ \mathcal{R}(\bm{a}_n) | \bm{a}_n \sim G([\theta_1^{j}, 
                \cdots, \theta_{i-1}^{j}, $
                \State $\qquad \qquad z,  \theta_{i+1}^{j-1}, \cdots, \theta_M^{j-1}]^T ) \}_{n=1}^{N}.$
                \State \Comment{Generate synthetic dataset}
                \State Score =  $\mbox{FID} (D^s, D^r)$ \Comment{FID calculation}
                \If {Score $<$ Optimal}
                    \State Optimal $=$ Score \Comment{Update FID}
                    \State $\theta_i^{j} = z$ \Comment{Update parameters}
                \EndIf 
            \EndFor
        \EndFor
    \EndFor
  \end{algorithmic}  
  \label{algorithm:attr_des}
\end{algorithm}  

Our loss function is parameterized by $\bm{\theta}$ which encodes the distribution of the configurable attributes. Following coordinate descent, we propose attribute descent (Alg. \ref{algorithm:attr_des}) to iteratively optimize each parameter. Specifically, we view each parameter as a coordinate in coordinate descent. In each iteration, we successively vary the value of a parameter so as to move toward a local minimum of the objective function.

Formally, to iteratively optimize the objective function (Eq.~\ref{eq:loss}) with respect to $\bm{\theta}$, we first initialize $\bm{\theta}$ at epoch $0$:
\begin{equation}
\bm{\theta}^{0} = [\theta_{1}^{0}, \cdots, \theta_{M}^{0}]^T.
\end{equation}
Then at the $j$th epoch, we get $\bm{\theta}^{j}$ from $\bm{\theta}^{j-1}$ by iteratively solving the single variable optimization problems. Specifically, at $i$th iteration in epoch $j$, we optimize a single parameter $\theta_{i}^{j}$ in $\bm{\theta}^{j}$, to its best value $z$ in the search space $\bm{S_{i}}$:
\begin{equation}
\begin{split}
\theta_{i}^{j} = \mathop{\arg\min}_{z \in \bm{S_{i}} } \mbox{FID}(D^s, D^r), 
\label{eq:theta}
\end{split}
\end{equation}
where 
\begin{equation}
\begin{split}
D^s = \{ \mathcal{R}(\bm{a}_n) |  \bm{a}_n \sim G([\theta_1^{j}, 
\cdots, \theta_{i-1}^{j},  \\z,  \theta_{i+1}^{j-1}, \cdots, \theta_M^{j-1}]^T ) \}_{n=1}^{N}.
\label{eq:D_A}
\end{split}
\end{equation}
In Eq.~\ref{eq:theta} and Eq.~\ref{eq:D_A}, an iteration is defined as the duration for which a single parameter $\theta_i^j, i=1,...,M$ undergoes an optimization process (from Step 7 to Step 14 in Alg. \ref{algorithm:attr_des}). An epoch is defined as the duration for which all parameters undergo one attribute descent round.

In attribute descent, each iteration performs a greedy search for a single parameter while values of the other parameters are fixed. Therefore, each iteration finds the best value for a single parameter, and an epoch gives values for the entire parameter vector $\bm \theta$. In our experiments, the entire optimization process usually converges in 2 epochs.

\begin{figure*}[t] 
    \centering
    \begin{center}  
        \includegraphics[width=1\linewidth]{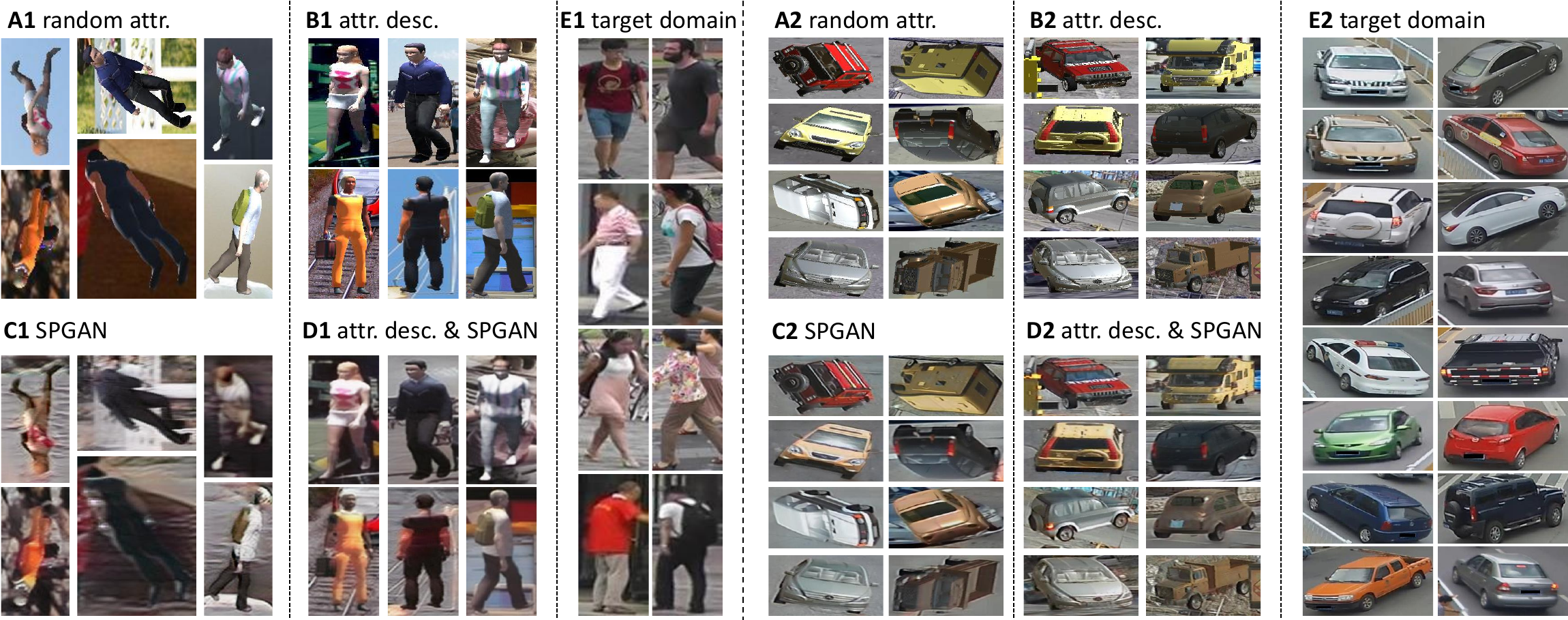}
        \caption{Examples of synthesized images for person re-ID \textbf{(left)} and vehicle re-ID \textbf{(right)} . \textbf{A1} and \textbf{A2}: images simulated by random attributes. \textbf{B1} and \textbf{B2}: image simulated by attributes optimized through attribute descent, to approximate the distribution of real-world target data \textbf{E1} and \textbf{E2}, respectively. \textbf{C1} and \textbf{C2}: we apply SPGAN \cite{deng2018similarity} to translate images in \textbf{A1} and \textbf{A2}, respectively into the style of the target domain. \textbf{D1} and \textbf{D2}: SPGAN is applied to images in \textbf{B1} and \textbf{B2}, respectively. We can observe that attribute descent changes the image content such as camera viewpoint and object orientation, while SPGAN amends image styles. The two forces are complementary to each other, as shown in our experiment. 
        }
        \label{fig:sample_images}
    \end{center}
    % \vspace{-1em}
\end{figure*}

\textbf{Convergence characteristics.} Attribute descent has the following properties in regarding model convergence.  

\emph{Fast conditional updates}. The problem configuration allows us to individually optimize parameters quickly. As stated in Section \ref{sec:attribute_descent}, we can update each attribute by using discrete values within a range. For example, the search space for azimuth is between $0^{\circ}$ and $330^{\circ}$ with a $30^{\circ}$ interval. Therefore, only 12 iterations are necessary, thus greatly decreasing the computational time. Note that, using attribute descent for complex tasks such as semantic segmentation would be result in a much slower overall procedure because these tasks have many more attributes to optimize. 

\emph{Strong stability.} Attribute descent can steadily converge because of its ``one-at-a-time'' optimization characteristics. That is, it gradually and greedily updates each parameter if and only if this update can improve the objective function. Compared with ``all-at-once'' algorithms like gradient descent which requires expert-level training skills to converge, attribute descent is easier to train and more stable in convergence.

%\red{\emph{We require attributes to be relatively independent}. If the optimal value of one attribute is completely independent of the other attributes, one round of attribute descent will lead to the optimal solution (assuming that each coordinate update finds the best value). To the other extreme, if the optimal value of an attribute is highly dependent on another attribute, attribute descent could converge more slowly. For attributes used in this work, we study their independence in Section~\ref{sec:quan_eval_ad}. }

\textbf{Analysis and visualization of the attribute descent process.} In Fig. \ref{fig:training}, we present how  %show the feasibility of using GAN evaluation metric to guide content domain adaptation in Figure~\ref{fig:training}. In particular, we shows how the 
task performance (re-ID accuracy) and domain discrepancy metric (FID) change during the attribute descent process. We observe that attributes are successively optimized when the FID value decreases and the re-ID accuracy mAP increases. Specifically, from the slope of the curves in Fig.~\ref{fig:training} A, when optimizing attributes in the order of ``orientation -> lighting -> camera pose'', we observe that after orientation attributes (\ie in-plane rotation and azimuth) are optimized, a large decrease in FID occurs from 147.85 to 91.14 and a large mAP increase occurs from 12.1\% to 21.94\%. Subsequently, after lighting attributes are optimized, we have -7.2 FID and +10.7\% mAP. The optimization of the camera attributes leads to -4.11 FID and +2.4\% mAP. These observations illustrate that all the attributes are useful for improving the training data quality. In addition, in Fig.~\ref{fig:training} B, the content of generated images becomes increasingly similar to the target real images through the optimization procedure, thus suggesting the effectiveness of attribute descent.

\subsection{Application Scenarios}% and Implementation Details}
\subsubsection{Training with Synthetic Data Only}

\textbf{Setting.} In the first application, we use generated synthetic data to replace real-world data for task model training. Specifically, we perform attribute descent to optimize synthetic data toward target data without labels, train task model on the adapted synthetic data, and then test the models on target test sets. 

\textbf{Obtaining target class or camera labels.} As stated in Section~\ref{sec:att_model}, attribute descent requires known class or camera labels, because we observe different attribute distributions for different categories or cameras; examples are shown in Section~\ref{sec:vis_content_bias}.

%For example, in collecting object classification datasets, \eg taking pictures using handheld cameras, we usually film persons or vehicles from a horizontal level, but film a dog more likely from a bird's eye view, thereby leading to viewpoint bias in different classes. Similarly, in collecting re-ID datasets, \eg taking vehicle images from city surveillance cameras, we usually have bi-modal vehicle orientations on a normal two-way, and have more diverse orientation in an intersection.  
%Thus, we optimize the attribute list for each category. 
%Thus, in this application,
%the major difficulty is how to 
In mining class or camera labels from the unlabeled target domain, we leverage pseudo labels for classification. Specifically, we use the model trained on synthetic data (\ie ObjectX) for pseudo label assignment with random attributes. Table \ref{tab:objectx2visda} shows that the classification model trained on ObjectX achieves 65.0\% average top-1 accuracy on the VisDA test set; therefore, the pseudo labels are relatively reliable. After we obtain the pseudo labels for the target domain, we perform attribute descent for each class by following Alg.~\ref{algorithm:attr_des}. Because ObjectX includes 7 classes, we optimize 7 attribute lists for the classification task.

For object re-ID, we simply perform attribute descent against each camera in the target domain, in order to simulate images with similar content to those from each camera. For example, we optimize 6 and 20 attribute lists for the Market and VeRi (both are target domains) training sets which have 6 and 20 cameras, respectively. Note that assuming knowledge of the camera label is a common practice in unsupervised domain adaptive re-ID \cite{zhong2019camstyle}.

%and perform classification on each part.the default part partition strategy for person re-ID. For person re-ID, we use original setting which vertically cut the image into six equal parts and perform classification on each part. In training IDE and PCB model, we use the ResNet-50 backbone. 
%For vehicle re-ID, we horizontally divide the picture into six equal parts. To achieve the state of the art accuracy in person re-ID, we adopt a third baseline method use the transformer \cite{he2021transreid}. Likewise, 
%to get the state-of-the-art accuracy in CityFlow, we use the DenseNet-101 \cite{huang2017densely} backbone and a setting from~\cite{luo2019bag} that combine of the cross-entropy loss and the triplet loss. 
% We use the ResNet-50 backbone for all the re-ID experiment except for training CityFlow. 

\subsubsection{Augmenting Target Training Data}
\label{sec:aug_training_data}

\textbf{Setting.} For this application, we use synthetic data to augment real-world training data. 
%, in purpose of achieving the higher accuracy. 
Specifically, labels of the target data are provided in both attribute descent and task model training. We combine the adapted synthetic data and real-world data and perform two-stage joint training~\cite{zheng2019vehiclenet} to obtain task models.% with the state-of-the-art accuracy.
For the classification task, when labels of the target training set are given, we optimize attributes for each category directly. Likewise, for the re-ID task, when labels for each camera are given, we optimize attributes for each camera.

\textbf{Two-stage training}~\cite{zheng2019vehiclenet} is conducted in training data augmentation wherein synthetic and real-world data are both used in training. We mix synthetic data and real-world data in the first stage and finetune with real-world data only in the second stage. In CityFlow, for example, in the first stage, we train on both real and synthetic data, where we classify vehicle images into one of the 1,695 (333 real + 1,362 synthetic) identities. In the second stage, we replace the classification layer with a new classifier fine-tuned on the real dataset (333 classes). When conducting the second stage training, we have a lower learning rate than that in the first stage, with details following \cite{zheng2019vehiclenet}. %in a purpose of finetune the task network to get higher accuracy. 

\subsubsection{Understanding Dataset Content Numerically}

Attribute descent provides a numerical way to understand (and sometimes visualize) the content of datasets. Specifically, given a certain category (or camera), we use attribute descent and synthetic models to build a proxy set, which has similar content distribution. After optimization, we can obtain the values of an attribute of interest for each image, and collectively obtain the value distribution for a dataset. We can then use either statistics or visualization tools to understand a certain aspect of the content of a dataset through the attribute. % distribution of , which is then plotted in some manner. %Then, we plot its value distribution in some manner. 
For example, we can visualize the viewpoint distribution in a 3D sphere (as shown in Section \ref{sec:vis_content_bias}). We can also use histograms to present the distribution of lighting intensity of a dataset. %This functionality of attribute descent allows us to understand dataset content and bias in a numerical way. 

%\red{To be shown in Section \ref{sec:vis_content_bias}, we visualize the viewpoint difference between different categories in object classification dataset, as well as viewpoint difference between different cameras in re-ID datasets. These patterns in classes or cameras reflect the viewpoint bias of datasets. For example, in Fig.~\ref{figure:dataset_bias_vid}, we show optimized viewpoint distribution for VehicleID is bi-modal. And for class \emph{knife} in VisDA, it has various in-plane rotations. These distinct characters of viewpoint form dataset content bias. } 

% optimized attributes a 

%Specifically, given an dataset, we use attribute descent to find the value of a certain attribute of this image. Then, for all images from a certain category (or camera), we plot the distribution of this attribute in some manner. In this section, we use attribute descent to demonstrate the \textbf{viewpoint bias} of VehicleID and VisDA-knife in Fig. \ref{figure:dataset_bias_vid}, and of the VisDA, Market, Duke, VeRi and CityFlow datasets in Fig. \ref{figure:content_bias_vis}. 
 
% \red{pls fill this part.}

\section{Experiment}

We evaluate the effectiveness of attribute descent on
%on three object-centric tasks: 
image classification, person re-ID and vehicle re-ID. %Image recognition aims to distinguish different classes, while object re-ID differentiates object IDs. 
In all tasks, given target data (Section \ref{sec:S_&_T_datasets}), we use attribute descent to synthesize a training set that has similar attribute distributions. % We show 

\subsection{Source and Target Datasets}
\label{sec:S_&_T_datasets}

\textbf{Image classification.} We use ObjectX (described in Section \ref{sec:3D_Asset-Acquisition}) as the source and the {VisDA}~\cite{peng2017visda} target set as the target domain. The original VisDA target set has 12 classes of real-world images. Among the 12 classes, we select 7 that are also included in ShapeNet V2~\cite{chang2015shapenet} and thus ObjectX. A total of 33,125 images are present in the 7 classes. For each class, the ratio of the number of training images to that of testing images is 1:7 or 1:1.

\textbf{Person re-ID.} We use PersonX as a source, and use two real-world datasets as target: Market-1501 (denoted as Market)~\cite{zheng2015scalable} and DukeMTMC-reID (denoted as Duke)~\cite{ristani2016MTMC}. Market has 1,501 IDs, 12,936 training images and 19,732 gallery images filmed by 6 cameras. A totoal of 751 out of 1,501 IDs are used for training and the remaining 750 are used for testing. The query set includes 3,368 bounding boxes from 750 identities. Duke contains 1,404 IDs and 36,441 images captured by 8 cameras. There are 16,522 images from 702 identities for training, 2,228 query images from another 702 identities and 17,661 gallery images for testing.

\textbf{Vehicle re-ID.} Apart from the synthetic VehicleX dataset, we use three real-world vehicle re-ID datasets as target domain data. VehicleID~\cite{liu2016deep} contains 222,629 images of 26,328 identities. Half the identities are used for training, and the other half are used for testing. Three test splits exists: ``Small'', ``Medium'' and ``Large'', representing the number of vehicles in the test set. Specifically, ``Small'' has 800 vehicles and 7,332 images, ``Medium'' has 1,600 vehicles and 12,995 images, and ``Large'' has 2,400 vehicles and 20,038 images. The VeRi-776 dataset~\cite{liu2016large} contains 49,357 images of 776 vehicles captured by 20 cameras. The vehicle viewpoints and illumination cover a diverse range. The training set has 37,778 images, corresponding to 576 identities; the test set has 11,579 images of 200 identities. There are 1,678 query images. The train / test sets share the same 20 cameras. We use ``VeRi'' for short in what follows. CityFlow-reID~\cite{tang2019cityflow} has more complex environments and has 40 cameras in a diverse environment where 34 of them are used in the training set. The dataset has in total 666 IDs where half are used for training and the rest for testing. We use ``CityFlow'' for short in the following context. 

\textbf{Evaluation protocol.} For image classification, we report the top-1 accuracy averaged over all the categories. For object re-ID, we use mean average precision (mAP) and cumulative match curve (CMC) scores to measure system accuracy, \eg ``Rank-1'' and ``Rank-5''. ``Rank-1'' denotes the success rate of finding the true match in the first rank, and ``Rank-5'' means the success rate of ranking at least one true match within the top 5. % rank-5 identification accuracy.

\begin{table*}[t]
\centering
\scriptsize
\caption{Comparing various training sets in object classification under synthetic training and data augmentation. We use \textbf{VisDA Target} as the target domain (real-world), and ObjectX (``OX'') as the source.  In terms of data composition, ``S'' represents synthetic data only, ``R'' denotes real-world data only, and ``R+S'' means both synthetic data and real-world data are used. Two validation-test splits are used for data augmentation, \ie 1:1 and 1:7. } 
\label{tab:objectx2visda}
\setlength{\tabcolsep}{2mm}
\begin{tabular}{l|l|c|c|l|ccccccc|c} 
\Xhline{1.2pt}
Application                     & Training data                      & Type               & \multicolumn{1}{l|}{Val-test split} & Model                     & plane                     & bus                       & car     & knife                     & mcycl                     & sktbrd                    & train & per-class  \\ 
\hline
\multirow{9}{*}{\begin{tabular}[c]{@{}l@{}}Synthetic\\Training\end{tabular}}  & VisDA Source                       & S                  & 1:7                                 & ResNet-50                  & 72.13                     & 38.57                     & 72.40   & 5.20                      & 90.41                     & 28.67                     & 84.51 & 56.0       \\ 
\cline{2-13}
                                                                              & \multirow{4}{*}{OX (Ran. Attr.)~}  & \multirow{4}{*}{S} & \multirow{4}{*}{1:7}                & ResNet-50                  & 77.96                     & 48.06                     & 62.33   & 40.88                     & 92.11                     & 57.99                     & 75.63 & 65.0       \\
                                                                              &                                    &                    &                                     & DAN~\cite{long2015learning}                       & 81.54                     & 47.40                     & 64.98   & 53.33                     & 80.16                     & 63.36                     & 79.38 & 67.2       \\
                                                                              &                                    &                    &                                     & ADDA~\cite{tzeng2017adversarial}                      & \multicolumn{1}{l}{83.01} & \multicolumn{1}{l}{51.01} & 60.11   & \multicolumn{1}{l}{55.65} & \multicolumn{1}{l}{69.63} & \multicolumn{1}{l}{71.03} & 71.91 & 66.0       \\
                                                                              &                                    &                    &                                     & SHOT~\cite{liang2020we}                      & \multicolumn{1}{l}{88.43} & \multicolumn{1}{l}{64.98} & 67.24   & \multicolumn{1}{l}{66.23} & \multicolumn{1}{l}{85.35} & \multicolumn{1}{l}{78.05} & 75.66 & 72.4       \\ 
\cline{2-13}
                                                                              & \multirow{4}{*}{OX (Attr. Desc.)~} & \multirow{4}{*}{S} & \multirow{4}{*}{1:7}                & ResNet50                  & \multicolumn{1}{l}{83.82} & \multicolumn{1}{l}{68.88} & 62.20   & \multicolumn{1}{l}{64.90} & \multicolumn{1}{l}{86.57} & \multicolumn{1}{l}{38.65} & 92.96 & 71.1       \\
                                                                              &                                    &                    &                                     & DAN~\cite{long2015learning}                    & \multicolumn{1}{l}{86.93} & \multicolumn{1}{l}{67.68} & 69.02   & \multicolumn{1}{l}{46.50} & \multicolumn{1}{l}{84.50} & \multicolumn{1}{l}{56.39} & 92.66 & 72.0       \\
                                                                              &                                    &                    &                                     & ADDA~\cite{tzeng2017adversarial}                      & \multicolumn{1}{l}{91.88} & \multicolumn{1}{l}{69.00} & 70.26   & \multicolumn{1}{l}{34.38} & \multicolumn{1}{l}{83.67} & \multicolumn{1}{l}{64.81} & 90.69 & 72.1       \\
                                                                              &                                    &                    &                                     & SHOT~\cite{liang2020we}                      & \multicolumn{1}{l}{93.92} & \multicolumn{1}{l}{78.26} & 75.09~~ & \multicolumn{1}{l}{42.20} & \multicolumn{1}{l}{91.13} & \multicolumn{1}{l}{78.35} & 88.83 & 78.3       \\ 
\hline\hline
\multirow{6}{*}{\begin{tabular}[c]{@{}l@{}}Data \\ Augmentation\end{tabular}} & VisDA Target                       & R                  & \multirow{3}{*}{1:7}                & \multirow{3}{*}{ResNet-50} & 96.24                     & 85.69                     & 91.47   & 94.33                     & 92.61                     & 90.88                     & 90.31 & 91.6       \\
                                                                              & VisDA Target+OX (Ran. Attr.)       & R+S                &                                     &                           & 94.89                     & 87.18                     & 93.55   & 93.39                     & 93.77                     & 91.03                     & 88.69 & 91.8       \\
                                                                              & VisDA Target+OX (Attr. Desc.)      & R+S                &                                     &                           & 96.90                     & 89.86                     & 94.33   & 95.04                     & 94.28                     & 92.38                     & 92.07 & 93.6       \\ 
\cline{2-13}
                                                                              & VisDA Target                       & R                  & \multirow{3}{*}{1:1}                & \multirow{3}{*}{ResNet-50} & 98.13                     & 89.38                     & 95.96   & 96.82                     & 95.34                     & 95.44                     & 93.39 & 94.9       \\
                                                                              & VisDA Target+OX (Ran. Attr.)       & R+S                &                                     &                           & 96.02                     & 86.74                     & 94.24   & 93.83                     & 93.47                     & 90.53                     & 89.07 & 92.0       \\
                                                                              & VisDA Target+OX (Attr. Desc.)      & R+S                &                                     &                           & 98.08                     & 91.56                     & 95.90   & 96.24                     & 96.00                     & 96.00                     & 94.48 & 95.5       \\
\Xhline{1.2pt}
\end{tabular}
\end{table*}

\begin{table}
\centering
\scriptsize
\caption{Comparison of various training sets in person reID under synthetic training and data augmentation. We use \textbf{Market} as the target and use PersonX (``PX'') as the source. A few state-of-the-art re-ID models are used. Data type notations are the same as those in Table \ref{tab:objectx2visda}. mAP (\%) and CMC scores (\%) are reported.} 
\label{tab:Resultspersonx2market}
\setlength{\tabcolsep}{0.7mm}
\begin{tabular}{l|l|c|c|ccc} 
\Xhline{1.2pt}
Appl.                         & Training data                 & Type                 & Model                & Rank-1         & Rank-5         & mAP             \\ 
\hline
\multirow{5}{*}{\begin{tabular}[c]{@{}l@{}}Syn. \\Training\end{tabular}} & ImageNet                & \multirow{3}{*}{R}   & \multirow{3}{*}{IDE~\cite{zheng2016mars}} & 6.38           & 14.55          & 1.92            \\
                                                                         & Duke                    &                      &                      & 42.31          & 61.88          & 18.07           \\
                                                                         & MSMT                    &                      &                      & 41.98          & 61.67          & 20.46           \\ 
\cline{2-7}
                                                                         & PX (Ran. Attr.)         & \multirow{2}{*}{S}   & \multirow{2}{*}{IDE~\cite{zheng2016mars}} & 17.01          & 33.49          & 6.30            \\
                                                                         & PX (Attr. Desc.)        &                      &                      & 34.71          & 51.60          & 15.01           \\ 
\hline\hline
\multirow{8}{*}{\begin{tabular}[c]{@{}l@{}}Data \\Aug.\end{tabular}}     & \multirow{4}{*}{Market} & \multirow{4}{*}{R}   & IDE~\cite{zheng2016mars}                  & 85.30          & 93.82          & 67.84           \\
                                                                         &                         &                      & PCB~\cite{sun2018beyond}                  & 92.49          & 96.85          & 76.67           \\
                                                                         &                         &                      & CBN~\cite{zhuang2020rethinking}                  & 94.35          & 97.91          & 83.63           \\
                                                                         &                         &                      & TransReid~\cite{he2021transreid}            & 94.72          & 98.47          & 88.03           \\ 
\cline{2-7}
                                                                         & Mar.+PX (Ran. Attr.)    & \multirow{4}{*}{R+S} & IDE~\cite{zheng2016mars}                  & 84.62          & 94.30          & 67.56           \\
                                                                         & Mar.+PX (Attr. Desc.)   &                      & IDE~\cite{zheng2016mars}                   & 87.23          & 94.60          & 71.17            \\
                                                                         & Mar.+PX (Attr. Desc.)   &                      & PCB~\cite{sun2018beyond}                  & 92.58          & 97.24          & 79.99           \\
                                                                         & Mar.+PX (Attr. Desc.)   &                      & TransReid~\cite{he2021transreid}             & \textbf{95.24} & \textbf{98.57} & \textbf{88.76}  \\
\Xhline{1.2pt}
\end{tabular}
\end{table} 

\begin{table}
\scriptsize
\centering
\caption{Comparison of various training set when using \textbf{Duke} as the target domain. Other setting and evaluation metrics are identical to Table~\ref{tab:Resultspersonx2market}. } 
\label{tab:Resultspersonx2duke}
\setlength{\tabcolsep}{0.7mm}
\begin{tabular}{l|l|c|c|ccc} 
\Xhline{1.2pt}
Appl.                        & Training data                & Type                 & Model                & Rank-1 & Rank-5 & mAP                        \\ 
\hline
                                                                        & ImageNet               & \multirow{3}{*}{R}   & \multirow{3}{*}{IDE~\cite{zheng2016mars}} & 4.76   & 11.09  & 1.63                       \\
\multirow{4}{*}{\begin{tabular}[c]{@{}l@{}}Syn.\\Training\end{tabular}} & Market                 &                      &                      & 32.63  & 47.94  & 17.39                      \\
                                                                        & MSMT                   &                      &                      & 46.50  & 64.59  & 28.04                      \\ 
\cline{2-7}
                                                                        & PX (Ran. Attr.)        & \multirow{2}{*}{S}   & \multirow{2}{*}{IDE~\cite{zheng2016mars}} & 22.17  & 38.96  & 10.09                      \\
                                                                        & PX (Attr. Desc.)       &                      &                      & 30.83  & 47.80  & 15.91                      \\ 
\hline\hline
\multirow{8}{*}{\begin{tabular}[c]{@{}l@{}}Data \\Aug.\end{tabular}}    & \multirow{4}{*}{Duke~} & \multirow{4}{*}{R}   & IDE~\cite{zheng2016mars}                  & 78.14  & 88.29  & 58.93                      \\
                                                                        &                        &                      & PCB~\cite{sun2018beyond}                  & 82.99  & 90.53  & 67.47                      \\
                                                                        &                        &                      & CBN~\cite{zhuang2020rethinking}                  & 84.82  & 92.51  & \multicolumn{1}{l}{70.13}  \\
                                                                        &                        &                      & TransReid~\cite{he2021transreid}            & 89.90  & 95.74  & 81.23                      \\ 
\cline{2-7}
                                                                        & Duke+PX (Ran. Attr.)   & \multirow{4}{*}{R+S} & IDE~\cite{zheng2016mars}                  & 75.99  & 87.70  & 55.77                      \\
                                                                        & Duke+PX (Attr. Desc.)  &                      & IDE~\cite{zheng2016mars}                  & 78.28  & 89.27  & 59.10                      \\
                                                                        & Duke+PX (Attr. Desc.)  &                      & PCB~\cite{sun2018beyond}                  & 84.25  & 92.15  & 70.71                      \\
                                                                        & Duke+PX (Attr. Desc.)  &                      & TransReid~\cite{he2021transreid}            & \textbf{90.41}  & \textbf{96.03}  & \textbf{81.43}                      \\
\Xhline{1.2pt}
\end{tabular}
\end{table}

\subsection{Experimental Details}
\label{sec:Experimental_Details}

\textbf{Attribute descent settings.} As discussed in Section~\ref{sec:att_model}, we model the distribution of the 6 attributes using GMM. Specifically, When using GMM, we set the number of Gaussian components to 3, 6, 1, 1, 1, and 1 for in-plane rotation, azimuth, light intensity, light direction, camera height, and camera distance, respectively. Meanwhile, as mentioned in Section \ref{sec:att_model}, only means of the Gaussians are optimized, and initialized from the lowest value in the search space.  For each learnable mean value $\theta_i, i=1,...,M$ in $\bm \theta$, the search space is specified in the range defined in Section \ref{sec:Configurable_attr}, and their search steps are 12, 12, 10, 6, 10, and 5 for in-plane rotation, azimuth, light intensity, light direction, camera height, and camera distance, respectively.

The covariance matrices are diagonal matrices with pre-defined diagonal elements. Specifically, the diagonal elements are (10, 10, 10), (20, 20, 20, 20, 20, 20), (0.63), (7.07), (0.4) and (0.6) for in-plane rotation, azimuth, light intensity, light direction, camera height, and camera distance, respectively. In attribute descent, the number of epochs of attribute descent is 2, which usually leads to convergence.

\begin{table*}[t]
\scriptsize
\centering
\caption{Comparison of various training sets when VehicleX is the source domain and \textbf{VehicleID} is the target domain. We examine the use case of training data augmentation. Data type notations and evaluation metrics are the same as those in Table~\ref{tab:Resultspersonx2market}. ``Small'', ``Medium'' and ``Large'' refer to the three test splits of the VehicleID test set~\cite{liu2016deep}. ``VX'' denotes the VehicleX dataset. ``VID'' means the VehicleID dataset.} 
\setlength{\tabcolsep}{3mm}
\begin{tabular}{l|c|l|ccc|ccc|ccc} 
\Xhline{1.2pt}
\multicolumn{1}{l|}{\multirow{2}{*}{Training data}} & \multicolumn{1}{l|}{\multirow{2}{*}{Type}} & \multirow{2}{*}{Model}         &                & Small          &                &                & Medium         &                &                & Large          &                 \\ 
\cline{4-12}
                                                                   & \multicolumn{1}{l|}{}                      &                                & Rank-1         & Rank-5         & mAP            & Rank-1         & Rank-5         & mAP            & Rank-1         & Rank-5         & mAP             \\ 
\hline
\multirow{3}{*}{VID}                                               & \multirow{3}{*}{R}                                          & RAM~\cite{liu2018ram}                         & 75.2           & 91.5           & -              & 72.3           & 87.0           & -              & 67.7           & 84.5           & -               \\
                                                                   &                                           & AAVER~\cite{khorramshahi2019dual}                       & 74.69          & 93.82          & -              & 68.62          & 89.95          & -              & 63.54          & 85.64          & -               \\
                                                                   &                                           & GSTE~\cite{bai2018group}                       & 75.9           & 84.2           & 75.4           & 74.8           & 83.6           & 74.3           & 74.0           & 82.7           & 72.4            \\ 
\hline
VID                                                                & R                                          & \multirow{3}{*}{IDE~\cite{zheng2016mars}} & 77.35          & 90.28          & 83.10          & 75.24          & 87.45          & 80.73          & 72.78          & 85.56          & 78.51           \\
VID+VX (Ran. Attr.)                                                & R+S                                        &                                & 80.2           & 93.98          & 85.95          & 76.94          & 90.84          & 82.67          & 73.45          & 88.66          & 80.55           \\
VID+VX (Attr. Desc.)                                               & R+S                                        &                                & \textbf{81.50} & \textbf{94.85} & \textbf{87.33} & \textbf{77.62} & \textbf{92.20} & \textbf{83.88} & \textbf{74.87} & \textbf{89.90} & \textbf{81.35}  \\
\Xhline{1.2pt}
\end{tabular}
\label{tab:Resultvehiclex2vehicleID_supervised}
% \vspace{-1em}
\end{table*}

\begin{table}[t]
\scriptsize
\centering
\caption{Comparison of various training sets when \textbf{VeRi} is the target domain. Data type annotations and evaluation metrics are the same as those in previous tables. The IDE and PCB baseline models are evaluated.}%In addition to some state-of-the-art methods, we summarize the results on top of two baselines, \emph{i.e.,} IDE~\cite{zheng2016mars} and PCB~\cite{sun2018beyond}.}
\setlength{\tabcolsep}{0.9mm}
\begin{tabular}{l|l|c|l|ccc} 
\Xhline{1.2pt}
Appl.                          & Training data               & Type & Model                & Rank-1               & Rank-5               & mAP                   \\ 
\hline
\multirow{5}{*}{\begin{tabular}[c]{@{}l@{}}Syn.\\ Training\end{tabular}} & ImageNet              & \multirow{3}{*}{R}    & \multirow{3}{*}{IDE} & 30.57                & 47.85                & 8.19                  \\
                                                                         & VehicleID~\cite{liu2016deep}                   &     &                      & 59.24                & 71.16                & 20.32                 \\
                                                                         & Cityflow~\cite{tang2019cityflow}              &     &                      & 69.96                & 81.35                & 26.71                 \\ 
\cline{2-7}
                                                                         & VX (Ran. Attr.)       & \multirow{2}{*}{S}    & \multirow{2}{*}{IDE~\cite{zheng2016mars}} & 43.56                & 61.98                & 18.36                 \\
                                                                         & VX (Attr. Desc.)      &     &                      & 51.25                & 67.70                & 21.29                 \\ 
\hline\hline
\multirow{8}{*}{\begin{tabular}[c]{@{}l@{}}Data \\ Aug.\end{tabular}}    & \multirow{4}{*}{VeRi} & \multirow{4}{*}{R}    & IDE~\cite{zheng2016mars}                  & 92.73                & 95.99                & 66.54                 \\
                                                                       
                                                                         &                       &     & VANet~\cite{chu2019vehicle}              & 89.78                & 95.99                & 66.34                 \\
                                                                         &                       &     & AAVER~\cite{khorramshahi2019dual}            & 90.17                & 94.34                & 66.35                 \\ 
                                                                           &                       &     & PCB~\cite{sun2018beyond}                  & 94.04 & 98.21 &  72.04 \\
\cline{2-7}
                                                                         & VeRi+PAMTRI~\cite{tang2019pamtri}           &  \multirow{4}{*}{R+S}  & PAMTRI~\cite{tang2019pamtri}           & 92.86                & 96.97                & 71.88                 \\
                                                                        & VeRi+VX (Ran. Attr.)  &   & IDE~\cite{zheng2016mars}                  & 93.21                & 96.20                & 69.28                 \\
                                                                         & VeRi+VX (Attr. Desc.) &  & IDE~\cite{zheng2016mars}                  & 93.44                & 97.26                & 70.62                 \\
                                                                         & VeRi+VX (Attr. Desc.) &  & PCB~\cite{sun2018beyond}                   & \textbf{94.34}       & \textbf{97.91}       & \textbf{74.51}        \\
\Xhline{1.2pt}
\end{tabular}
\label{tab:Resultsvehiclex2veri}
% \vspace{-2.0em}
\end{table}

 \begin{table}[t]
 \scriptsize
\centering
\caption{Comparison of various training sets when \textbf{CityFlow} is the target domain. Data type annotations and evaluation metrics are the same as those in previous tables. We use IDE~\cite{zheng2016mars} as the task model with both the cross-entropy (CE) loss and triplet loss.}
\setlength{\tabcolsep}{0.9mm}
\begin{tabular}{l|c|l|ccc} 
\Xhline{1.2pt}
Training data                    & \multicolumn{1}{l|}{Type} & Model         & Rank-1         & Rank-20        & mAP             \\ 
\hline
\multirow{3}{*}{CityFlow~} & \multirow{3}{*}{R}        & BA~ ~         & 49.62          & 80.04          & 25.61           \\
                           &                           & BS~ ~         & 49.05          & 78.80          & 25.57           \\
                           &                           & IDE (CE+Tri.) & 56.75          & 72.24          & 30.21           \\ 
\hline 
CityFlow+PAMTRI~\cite{tang2019pamtri}         & \multirow{3}{*}{R+S}      & PAMTRI~\cite{tang2019pamtri}     & 59.7           & 80.13          & 33.81           \\
CityFlow+VX (Ran. Attr.)   &                           & IDE (CE+Tri.)~\cite{zheng2016mars} & 63.59          & 82.60          & 35.96           \\
CityFlow+VX (Attr. Desc.)  &                           & IDE (CE+Tri.)~\cite{zheng2016mars} & \textbf{64.07} & \textbf{83.27} & \textbf{37.16}  \\
\Xhline{1.2pt}
\end{tabular}
\label{tab:Resultsvehiclex2cityflow}
% \vspace{-2.0em}
\end{table}

\textbf{Image style transformation.} For object re-ID (both person and vehicle), we use SPGAN \cite{deng2018image} to transfer the appearance of synthetic data to that of the target domain. For image classification, we do not use style transformation. %After obtaining content-adapted data, we apply appearance-level style transformation for re-ID tasks. %based on CycleGAN~\cite{zhu2017unpaired} \red{for image classification}. S
%Specifically, we use SPGAN~\cite{deng2018image}, a widely used algorithm in identity-preserving image translation, where we use synthetic data with random attributes as the source domain and real data as the target domain. % style level domain adaptation. 
%Sample translation results are shown in Fig.~\ref{fig:sample_images}. %and influence is shown in Fig.~\red{xxx}. 
% In our implementation, each image is sized to 256 $\times$ 256. %In training image translation models from synthetic to real, 
 %For the purpose of quick experiment, 
%For optimized synthetic data, we directly deploy the translation model trained from random-attribute synthetic data and real-world data. Note that style transfer is only used in re-ID tasks for improving accuracy. In object classification task, we use content-adapted images directly.  
%We do not additionally train a translation model for transferring learned attributes data, but use model trained on random attributes directly. 

\textbf{Task model configuration.} For the classification task, we use ResNet-50 \cite{he2016deep} to classify the 7 classes. For person re-ID, we use multiple task models, including ID-discriminative embedding (IDE)~\cite{zheng2016mars}, the part-based convolution (PCB)~\cite{sun2018beyond}, and TransReid \cite{he2021transreid}. Note that when implementing these task models, we use their official implementations with default hyperparameters including learning rate and training epochs. %Our first model is the ID-discriminative embedding (IDE)~\cite{zheng2016mars}. 
For IDE, we adopt the strategy from~\cite{luo2019bag} which uses ResNet-50 \cite{he2016deep} and adds batch normalization and removes ReLU after the final feature layer. For PCB, we use the ResNet-50 backbone and vertically partitions an image into six equal horizontal parts. For vehicle re-ID, we also use IDE and PCB. For IDE, we also use the ResNet-50 backbone. For PCB, we use the ResNet-50 backbone and horizontally divide an image into six parts.

\subsection{Quantitative Evaluation of Attribute Descent}
\label{sec:quan_eval_ad}
Given a target set, attribute descent allows us to synthesize a dataset that has a similar distribution on the content (attribute) level. In this section, we demonstrate three application scenarios of the synthesized data in object-centric tasks: training with synthetic data only, real-synthetic data augmentation, and dataset bias visualization. In each application scenario, we compare attribute descent with several existing methods.

\textbf{Effectiveness of attribute descent for synthetic-data-only training.} After generating content-adapted synthetic data with attribute descent (visual examples in Fig. \ref{fig:sample_images}), we train the subsequent classification / re-ID models with generated synthetic data only. We mainly compare the optimized synthetic dataset with those generated with random attributes, a commonly used baseline in the community \cite{tremblay2018training, kim2017learning}. Here, ``random'' means that attributes follow the uniform distribution, where their value ranges are the same as the search space of learned attributes. %Besides, we also compare with existing real-world or synthetic training sets in these tasks. 
Experimental results on the VisDA target set, Market, Duke, VehicleID, VeRi and CityFlow are shown in Table \ref{tab:objectx2visda}, Table \ref{tab:Resultspersonx2market}, Table \ref{tab:Resultspersonx2duke}, Table \ref{tab:Resultvehiclex2vehicleID_unsupervised}, Table \ref{tab:Resultsvehiclex2veri} and Table \ref{tab:Resultsvehiclex2cityflow}, respectively. 

\begin{figure}[t]
\centering
\includegraphics[width=0.47\textwidth]{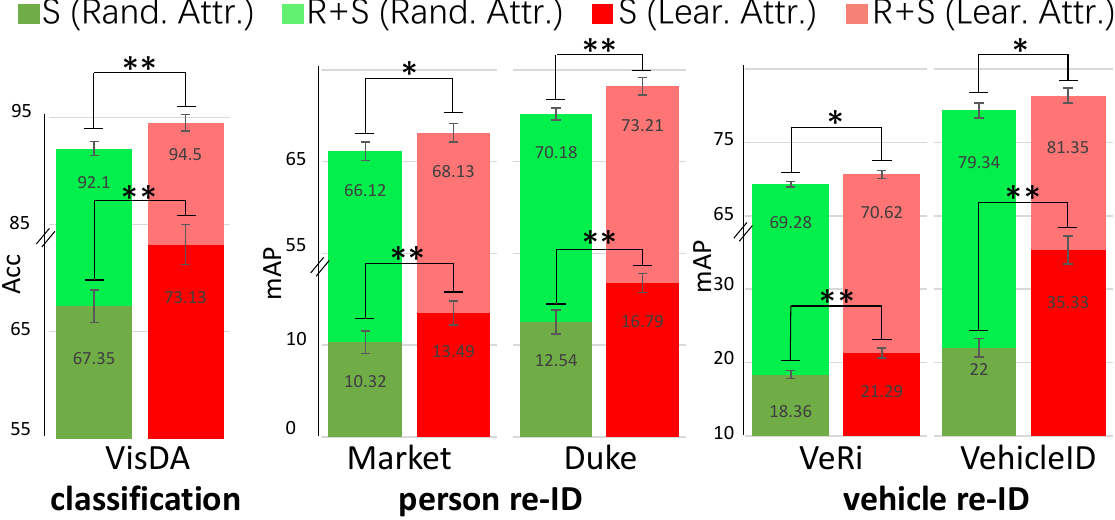}
\caption{Comparison of training sets synthesized from learned attributes and random attributes. Experiments are conducted on image classification (VisDA), person re-ID (Market and Duke), and vehicle re-ID (VeRi and VehicleID). Top-1 recognition accuracy (\%) and mAP (\%) are used. Two application scenarios are evaluated: training with synthetic data only (``S'') and training data augmentation (``R+S''). We clearly observe that learned attributes contribute to better training sets. Statistical significance analysis is performed, where  $*$ means statistically significant (\ie $0.01 < p$-value $< 0.05$) and $**$ denotes statistically very significant (\ie $p$-value $< 0.01)$. 
}
\label{figure: stas_analy}
% \vspace{-1.8em}
\end{figure}

From these results, we observe that when using only synthetic data for training, the data generated from learned attributes achieve much higher task accuracy than those generated from random attributes. For example, when adapting ObjectX to VisDA, attribute descent results in a +6.1\% improvement in per-class accuracy over using random attributes. When adapting PersonX to Market, attribute descent yields a +7.69\% improvement in Rank-1 accuracy over using random attributes. From VehicleX to VeRi, attribute descent again contributes to a +7.69\% improvement in Rank-1 accuracy. 

Of note, accuracy under this application scenario is usually lower than that of the state of the art or that produced by in-distribution training sets. This difference is understandable, because synthetic data have a relatively low resemblance with respect to the target data in terms of appearance.

%Besides, we find that the learned training set has mixed performance when compared with existing real/synthetic training data. On the one hand, when compared with in-distribution training data (\eg, Duke training set when using Duke as the test domain), synthetic training sets lead to inferior accuracy. This is understandable because synthetic data have relatively low appearance resemblance with the target data. On the other hand, when compared with

\textbf{Effectiveness of optimized synthetic data in augmenting the target training data.} After optimizing attributes to mimic the target domain, we mix the generated synthetic data with the target training data (with labels) to train the recognition / re-ID models. Apart from the two-stage training strategy (see Section~\ref{sec:aug_training_data}), no additional training skills are employed. We again compare attribute descent with random attributes. Experimental results for the three object-centric tasks are summarized across Table \ref{tab:objectx2visda}, Table \ref{tab:Resultspersonx2market}, Table \ref{tab:Resultspersonx2duke}, Table \ref{tab:Resultvehiclex2vehicleID_supervised}, Table \ref{tab:Resultsvehiclex2veri} and Table \ref{tab:Resultsvehiclex2cityflow}. Under this application, we also observe consistent improvement brought by the additional synthetic data. For example, from ObjectX to VisDA, the improvement in learned attributes over random attributes is +1.8\% in top-1 recognition accuracy. From PersonX to Market and from VehicleX to VeRi, the improvements in mAP are +2.61\% and +1.23\%, respectively. 
%Although 
The improvements appear numerically smaller than those in the ``training with synthetic only'' setting, 
% we show that the improvement is statistically significant (refer Fig.~\ref{figure: stas_analy}). In fact, 
%it is understandable that the improve in joint training is less than using synthetic data alone, 
because the latter sits on a relatively low baseline due to its \emph{appearance} discrepancy between source the target data. To summarize, the superiority of attribute descent over random attributes is also shown in Fig.~\ref{figure: stas_analy}.   %We note that the improvement of using synthetic data alone as a training set for task model is more significant than joint training. When the training set consists of only the synthetic data, a higher quality of attributes will have a more direct impact on the task performance. 

In image classification, we evaluate two ratios (\ie 1:1 and 1:7) of the number of training data to that of test data. Table \ref{tab:objectx2visda} indicates that our method brings consistent superiority to training with target data only and augmentation with randomly synthesized data under both ratios. When using a 1:1 ratio, the improvement is smaller in magnitude, a finding that is understandable because using more training data would lead to higher baseline accuracy.

In object re-ID, we also demonstrate the benefit of synthetic data augmentation to a few existing models. For example, in Table \ref{tab:Resultspersonx2market}, when using IDE, PCB and TransReid architectures, augmenting the training set with optimized synthetic data is consistently beneficial compared with using real data only. Similar observations are made on the PersonX to Duke setting (Table \ref{tab:Resultspersonx2duke}) and the VehicleX to VeRi setting (Table \ref{tab:Resultsvehiclex2veri}).

\textbf{Comparison with existing gradient-free methods.} We compared the proposed attribute descent with random search, evolutionary algorithm (\ie genetic algorithm), Bayesian optimization, and reinforcement learning (\ie LTS). These methods have been used as a strong baseline in hyper-parameter search and neural architecture search \cite{bergstra2012random}, and only reinforcement learning has been previously used for content-level domain adaptation \cite{ruiz2019learning}. 

\begin{figure}[t]
 \begin{minipage}[c]{0.27\textwidth}
    \includegraphics[width=\textwidth]{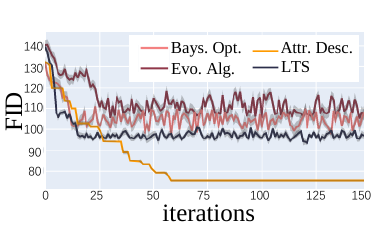}
  \end{minipage}
  % \vspace{-1mm}
  % \hfill
\begin{minipage}[c]{0.21\textwidth}
    \caption{Convergence compari-\\son between attribute descent and existing gradient-free methods including Bayesian optimization, evolutionary algorithm and LTS.  Gray regions show the error bar for loss curves. 
    }
  \label{fig:Convergence}
  \end{minipage}
  % \vspace{-4mm}
\end{figure}

Specifically, for the random search, we randomly sample attribute values 200 times and choose the attribute list with the best FID score. For the evolutionary algorithm, we use a generic algorithm with a fitness function equal to FID~\cite{xie2017genetic}. For Bayesian optimization, we use the pipeline stated in~\cite{shahriari2015taking}. For reinforcement learning, we reproduce the LTS structure~\cite{ruiz2019learning} and replace the task loss with the FID score. When comparing these methods, we use the same distribution definition and initialization as attribute descent. For a fair comparison, we report the best results after 200 iterations of training (\ie computing the FID score 200 times and taking the lowest FID score). Comparison results are shown in Fig.~\ref{fig:Convergence} and Table~\ref{tab:Resultvehiclex2vehicleID_unsupervised}, from which four observations are made.

\begin{figure}[t]
\centering
\includegraphics[width=0.47\textwidth]{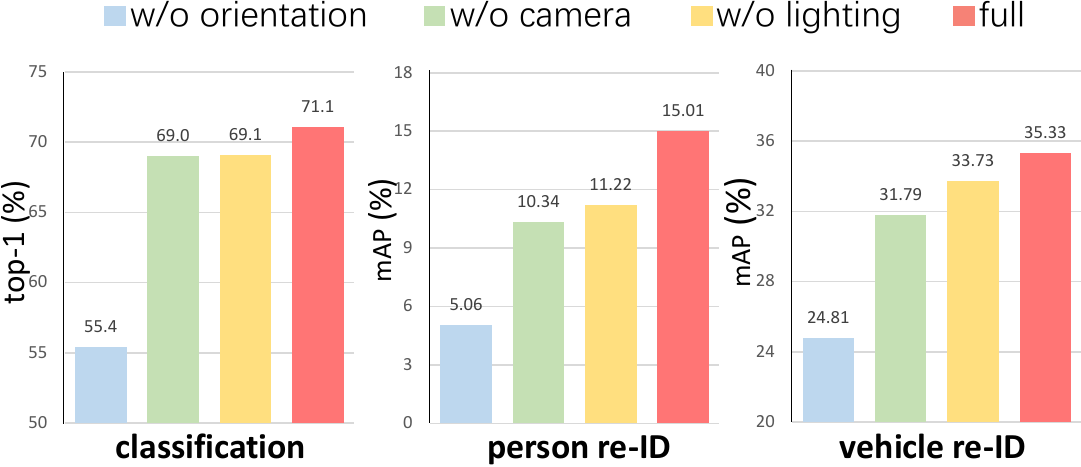}
\caption{Ablation studies of each group of attributes: object orientation, camera pose and lighting. Each ablation experiment leaves a certain group of attributes unoptimized (\ie following the uniform distribution) and is compared with the full system. Top-1 accuracy (\%), mAP (\%) and mAP (\%) are reported on image classification, person re-ID and vehicle re-ID tasks, respectively. We use VisDA, Market, and VehicleID as the target domain for the three tasks, respectively.}%classification, person re-ID and vehicle re-ID, respectively.  }
\label{figure:ablation}
% \vspace{-1.8em}
\end{figure}

First, under the same task network IDE~\cite{zheng2016mars}, learned attributes (regardless of which optimization method is used) outperform random attributes in both FID and mAP, demonstrating the benefit of attribute learning for alleviating content difference. Second, random search does not perform well in a limited search time. In fact, it is shown that random search is more effective when many unimportant parameters exist~\cite{bergstra2012random}. But in our search space, all the attributes significantly contribute to the distribution differences as shown in Fig.~\ref{fig:training} and Fig.~\ref{figure:ablation}. 
Third, the evolutionary algorithm, Bayesian optimization, and LTS appear to fall into an inferior local optimum and thus do not produce a lower FID score than attribute descent. To empirically understand such difference, we find that synthetic data optimized by LTS in mimicking the VehicleID dataset exhibit either the car front or rear, whereas VehicleID actually contains both car front and rear. In comparison, our method can sense both directions as it can iterate the entire search space. Fourth, shown in Fig.~\ref{fig:Convergence}, compared with existing gradient-free methods, attribute descent has the benefit of stable convergence due to its greedy search nature. Given these benefits, attribute descent presents itself as a straightforward yet effective baseline for syn2real content-level domain adaptation.

\textbf{Positioning among state-of-the-art systems.} This article aims to demonstrate the consistent improvement gained through the use of optimized synthetic data, instead of focusing on achieving a new state of the art. Nevertheless, the system augmented with learned synthetic data has very competitive accuracy. Comparisons with several representative state-of-the-art methods are summarized in Table \ref{tab:Resultspersonx2market}, Table \ref{tab:Resultspersonx2duke}, Table \ref{tab:Resultvehiclex2vehicleID_supervised}, Table \ref{tab:Resultsvehiclex2veri} and Table \ref{tab:Resultsvehiclex2cityflow}. For example, when Market is jointly trained with personX, our system outperforms TransReID \cite{he2021transreid} with real data only by +0.52\% in Rank-1 accuracy. Similarly, on Duke and VehicleID (small), our system exceeds TransReID \cite{he2021transreid} and GSTE \cite{bai2018group} by +0.51\% and +5.6\% in Rank-1 accuracy. On the VeRi-776 dataset, the mAP of our system using the PCB backbone is 74.51\%, which is +2.63\% higher than that using PAMTRI \cite{tang2019pamtri}.

\begin{figure}[t]
\centering
\includegraphics[width=0.47\textwidth]{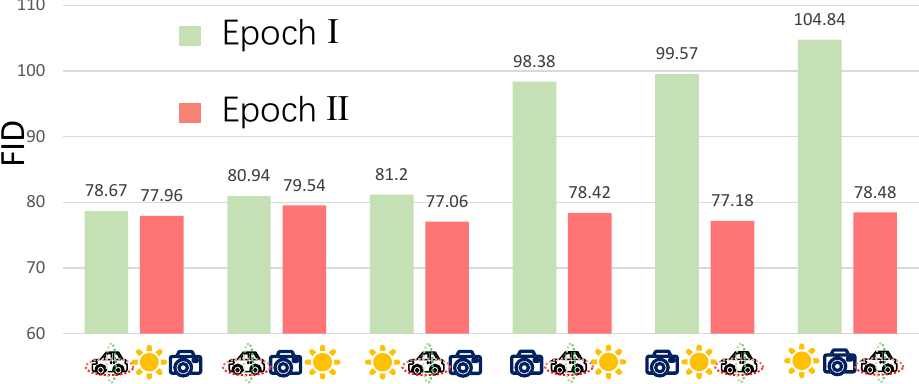}
\caption{Comparison of different attribute orders in attribute descent optimization. %Attribute correlation study. W
We show FID values between generated data and the VehicleID dataset after epoch I and II in attribute descent, which is performed for a total of two epochs (see Alg. \ref{alg:Framwork}). Various optimization orders of attributes are tested. %Study of attribute dependency in terms of distribution difference (FID). 
Each order is described by icons representing attributes of object orientation (azimuth and in-plane rotation), camera pose (camera height and distance) and lighting (light direction and intensity). 
%``C'', ``O'' and ``L'' refer to camera pose, object orientation and lighting, respectively. 
Under different attribute orders, we observe similar FID values after epoch II.}% It suggests 
%the FID values are generally similar, suggesting that 
%the correlation among attributes is weak; in other words, they are mostly independent with each other.} %, \emph{resp}.}
\label{figure:dependency}
% \vspace{-1.8em}
\end{figure}

\textbf{Impact of attribute order in attribute descent.} In Fig. \ref{figure:dependency}, we investigate the dependency among attributes by testing whether the order of attributes matters in attribute descent. Using different attribute orders in attribute descent optimization and comparing the FID scores between generated data and target data at epoch I and epoch II, yields two observations. First, after the first epoch, some orders generate lower FIDs than others. For example, the order ``orientation -> lighting -> camera pose'' results in lower FID than ``lighting -> camera pose -> orientation''. This difference is because orientation accounts more for the discrepancy between synthetic data and real data than camera pose and lighting. %If not being first optimized, the large discrepancy caused by orientation will lead less prominent supervision signals when optimizing lighting and camera pose. 
Second, although different orders may give different FID values after epoch I, their FID values (and accuracy, not shown in this figure) become similar after epoch II. This property is associated with the coordinate descent algorithm, wherein the order of coordinates in the optimization does not affect final performance. %In other words, the order of attributes in attributes descent does not affect the FID value and accuracy in Epoch II (we only use two epochs). 
\begin{table}[t]
\centering
\scriptsize
\caption{Comparison of attribute descent and existing gradient-free methods. The training set comparison when \textbf{VehicleX} is the source domain and \textbf{VehicleID} is the target domain, under training with synthetic data only. The IDE~\cite{zheng2016mars} model and the ``Large'' train-test split is used. Lower FID indicates lower domain discrepancy with VehicleID. } 
\label{tab:Resultvehiclex2vehicleID_unsupervised}
\setlength{\tabcolsep}{2.9mm}{
\begin{tabular}{l|c|cccc} 
\Xhline{1.2pt}
Training data          & \multicolumn{1}{l|}{Type} & FID    & Rank-1 & Rank-5 & mAP    \\ 
\hline
VeRi~            & \multirow{2}{*}{R}        & 45.39 & 28.00  & 41.11  & 38.59  \\
Cityflow~        &                           & 75.36 & 38.23  & 53.70  & 45.57  \\ 
\hline
VX (Ran. Attr.)  & \multirow{6}{*}{S}        & 134.75 & 18.76  & 30.11  & 22.00  \\
VX (Ran. Sear.)  &                           & 109.94 & 21.84  & 35.29  & 26.35  \\
VX (Evo. Algo.) &                           & 105.14 & 21.97  & 35.78  & 27.12  \\
VX (Bay. Opti.)  &                           & 99.64 & 22.57  & 38.15  & 30.05   \\
VX (LTS)~        &                           & 95.27  & 24.03  & 38.62  & 32.21  \\
VX (Attr. Desc.) &                           & 77.96  & 28.04  & 41.85  & 35.33  \\
\Xhline{1.2pt}
\end{tabular}
}
\end{table}

{\textbf{Impact of different attributes.} We perform ablation studies on each group of attributes: object orientation, camera pose and lighting. Results on the application of training with synthetic data only, and tasks of classification, person re-ID, and vehicle re-ID are summarized in Fig.~\ref{figure:ablation}. The results provide us with interesting insights regarding the importance of different attributes in these tasks. First, we observe that all three groups of attributes are necessary for good optimization results, where omitting any of them would decrease the accuracy. For example, when we use random values for orientation attributes, task accuracy drops by 15.7\% and 9.95\% in top-1 recognition rate and mAP on classification and person re-ID task, respectively. Second, we find that attributes have different importance. Specifically, orientation attributes are the most important. % much more important than camera pose, and both of them are more important than lighting. 
For example, in the vehicle re-ID task, not optimizing vehicle orientation leads to a -10.52\% drop in mAP, while is much more than the drop caused by omitting camera pose (-3.54\%) and lighting (-1.6\%).}

\begin{figure*}[t]
\centering
\includegraphics[width=1\textwidth]{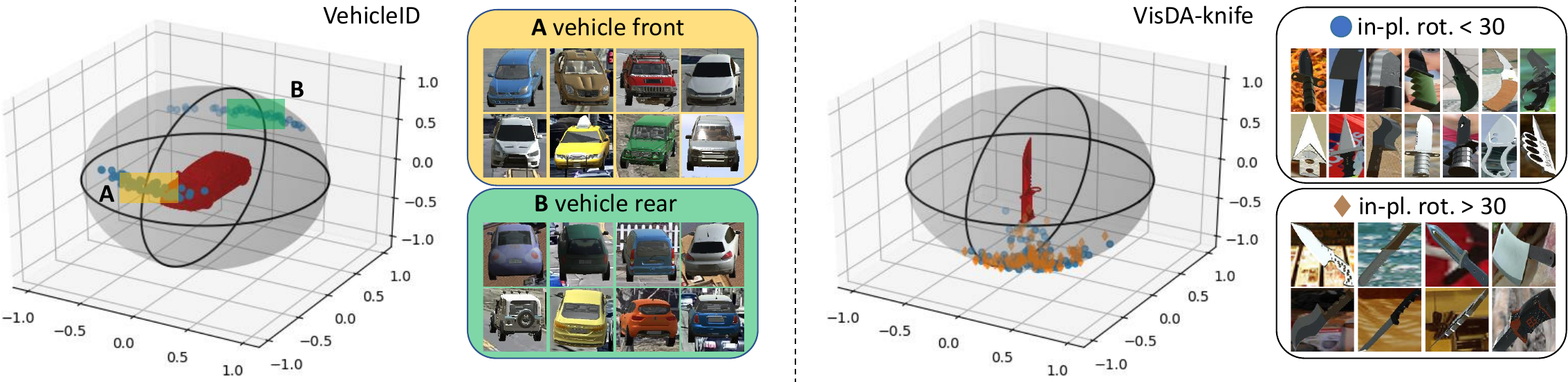}
\caption{Viewpoint distribution visualization for VehicleID and class \emph{knife} in VisDA. (\textbf{Left}:) by estimating the orientation parameters for each vehicle, our method shows that the viewpoint distribution on VehicleID is bi-modal, where orientations are concentrated in the \textbf{A} front or \textbf{B} rear. (\textbf{Right}:) our visualization method shows that the viewpoint of the class \emph{knife} on VisDA exhibits various in-plane rotations. The blue dots indicate image samples with in-plane rotation $< 30^{\circ}$ while the orange dots mean in-plane rotation $> 30^{\circ}$.
} 
\label{figure:dataset_bias_vid}
% \vspace{-1.8em}
\end{figure*}

% Second, we find that attributes have different importance for different tasks. For example, the orientation-related attributes are very important for vehicle re-ID and image classification, but less so for person re-ID. It is because vehicles and daily objects exhibit certain levels of orientation bias, which leads to deformed object shapes and thus affects the features in a non-trivial way. In comparison, while pedestrians also have orientation bias, different orientations do not significantly influence the shapes of pedestrians and thus have limited impact on the accuracy.

\begin{figure*}[t]
\centering
\includegraphics[width=1\textwidth]{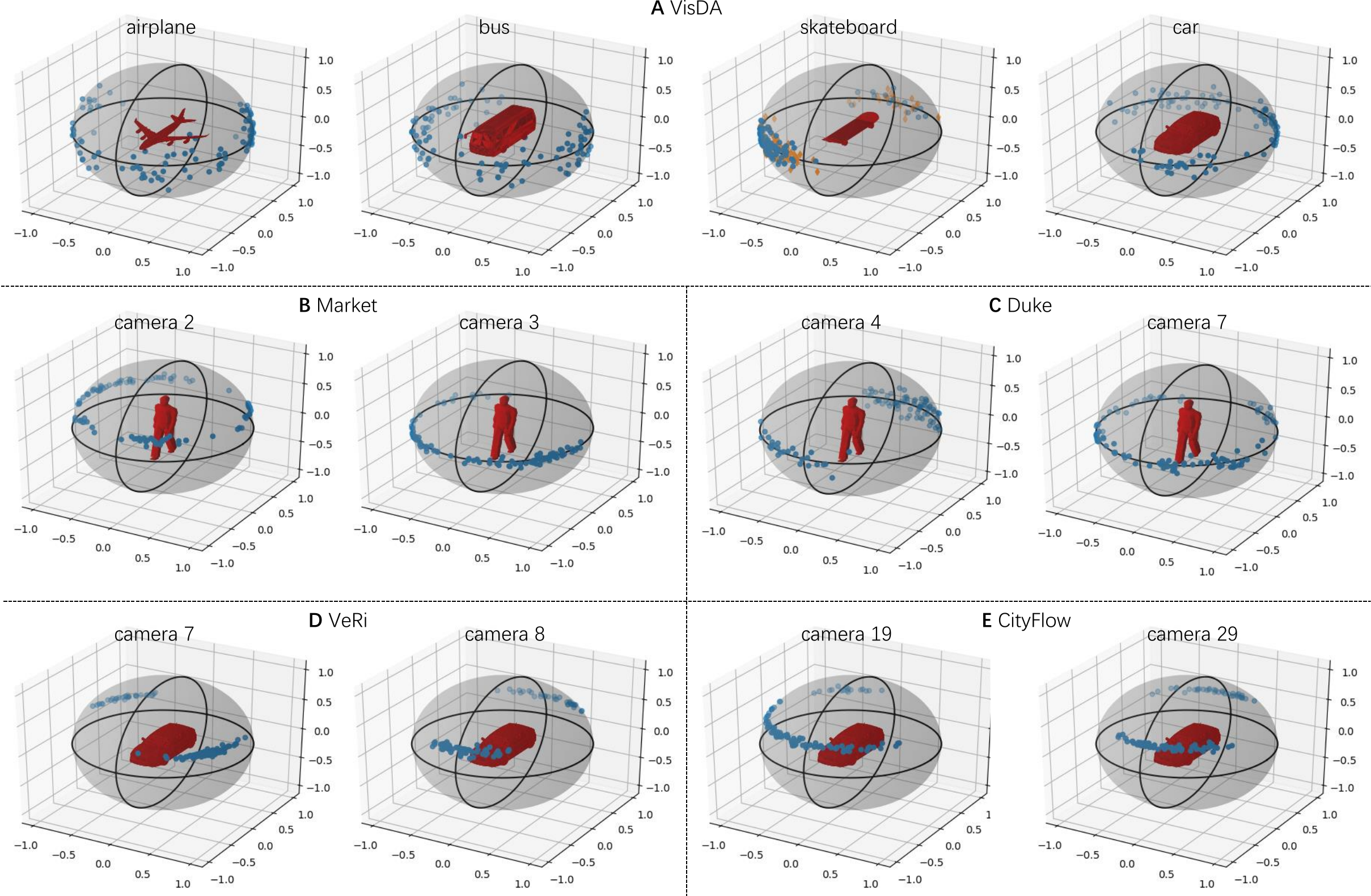}
\caption{Viewpoint distribution visualization for different categories in VisDA and different cameras in Market, Duke, VeRi and CityFlow. In image classification (VisDA), we observe that each object has a specific and unique view point pattern. Likewise, each camera in the re-ID task has a distinct pattern. These patterns in classes or cameras reflect the viewpoint bias of datasets. } %, \emph{resp}.}
\label{figure:content_bias_vis}
\vspace{-1em}
\end{figure*} 

\subsection{Numerically Understanding Dataset Content}
\label{sec:vis_content_bias}
This section uses \emph{viewpoint} as an example to showcase the application of attribute descent in numerically understanding dataset content. 
%, more specifically, viewpoint bias, between datasets. 
As shown in Fig.~\ref{figure:dataset_bias_vid} and Fig. \ref{figure:content_bias_vis}, after performing attribute descent on the corresponding synthetic assets, we plot obtained viewpoint value distributions on the unit sphere, where each point on the unit sphere represents a camera pointing toward the center of sphere.% and aligned object. 
The blue points show in-plane rotation of $< 30^{\circ}$. The orange points indicate in-plane rotation of $> 30^{\circ}$.

%In this section, we use attribute descent to visualize the \textbf{viewpoint bias} of VehicleID and VisDA-\emph{knife} in Fig. \ref{figure:dataset_bias_vid}. We then visualize Market, Duke, VeRi and CityFlow and some other classes of VisDA datasets in Fig. \ref{figure:content_bias_vis}. 

\textbf{Viewpoint distribution and bias for various classes in VisDA.} In Fig. \ref{figure:dataset_bias_vid} (right) and Fig. \ref{figure:content_bias_vis} \textbf{A}, we observe a significant viewpoint bias in the five categories. For example, we find that \emph{airplane}, \emph{car} and \emph{bus} are usually filmed vertically (in a normal erect position), because the in-plane rotation angles learned for the three class are usually less than 30 degrees. In contrast, \emph{knife} and \emph{skateboard} are often filmed from a certain oblique angle with significant in-plane rotations.   %but often film  and  from a certain amount of oblique angle on in-plane rotation. This is due to fact that airplane and bus are usually erect but knife and skateboard are not. 
Moreover, when capturing \emph{airplane}, \emph{bus} and \emph{skateboard} images, the camera is usually at the same height as the object, but for \emph{knife}, it is usually as either a higher or lower position. 

\textbf{Viewpoint distribution and bias for different cameras in re-ID datasets.} In Fig. \ref{figure:content_bias_vis} \textbf{B}-\textbf{E}, we observe very different viewpoint patterns of different cameras. 
%\textbf{A significant viewpoint bias for different cameras in real world re-ID datasets.} We visualize the viewpoint bias for  in figure , respectively. From these figures, we observe a significant viewpoint bias between cameras in a single dataset. 
On the Market dataset, the viewpoint distribution for camera 2 is distinct from that for camera 3. Specifically, we observe that camera 2 is higher than camera 3 and the azimuth of camera 2 covers a broader range than that of camera 3. Likewise, on the Duke dataset, camera 4 mainly films people from front or rear angles, whereas camera 7 films from nearly all directions of the azimuth. Similar phenomena are also observed in vehicle re-ID datasets. In VeRi, for example, in contrast to camera 8, which films only car fronts and car rears, camera 7 mainly films vehicles from the side. In summary, significant viewpoint bias exist for different cameras in re-ID datasets. Such bias comes from the fact that camera positions are usually fixed and that objects (person or vehicle) regularly follow predefined lanes. Bias among cameras inevitably leads to bias between datasets, and potentially decrease accuracy when deploying models. 

%\textbf{A significant viewpoint bias for different classes in a real world classification dataset.} Four subfigures in Figure \ref{figure:content_bias_vis} A visualize viewpoint bias for class airplane, bus, knife and skateboard respectively. Compared with viewpoint difference in re-ID task, more significant difference are found. Firstly, 

\section{Discussion} 
\label{sec:method_disscussion}

\textbf{Optimization under non-differentiable simulation functions.} Our system is non-differentiable because of the Unity rendering function. Under this circumstance, the gradient can be estimated by a few existing methods such as finite-difference~\cite{kar2019meta} and reinforcement learning~\cite{ruiz2019learning}. 
These methods are best applied in scenarios with a relatively large number of parameters (at least hundreds or thousands) to optimize. However, in object-centric tasks, far fewer attributes to optimize, so we instead propose a stable and efficient optimization approach.

\textbf{The relationship between distribution shift and task accuracy} has been examines in several recent studies. Deng \etal assume a fixed training set and quantitatively measure the strong negative correlation between accuracy and distribution shift of the test set in image classification~\cite{deng2021labels,deng2021does}. In comparison, we assume a fixed test set instead and use the negative correlation in method design: a better training set would have a smaller distribution shift from the test set. This assumption is verified in both object classification and re-ID, which complements \cite{deng2021labels,deng2021does} from both the assumption and application perspectives. 

\textbf{Can we use metrics other than FID to provide supervision signals?} Two other methods could potentially be used to measure distribution gaps: building a discriminator with an adversarial loss or training a task network with the task loss. However, in our preliminary experiment, the \textbf{discriminator} method is prone to detecting the large difference between synthetic and real data and thus continually tells the generator that its generated data have poor quality. This problem breaks the Nash equilibrium between the generator and discriminator, thus hindering us from obtaining an effective generator. On the other hand, \textbf{a task network} can provide accurate supervision signals but is infeasible when target domain labels are not provided. Furthermore, for object re-ID which is evaluated across cameras, the overall supervision provided by the task network does not reflect the training data quality in single cameras and thus poses difficulty in synthesizing data in each camera. As a result, we focus on the difference between features, using FID~\cite{heusel2017gans} to quantitatively measure the distribution difference between two datasets.

\textbf{Application scope.} To demonstrate the effectiveness of attribute descent, this article focuses on object-centric tasks, which are either fundamental or have very important applications in the real world, and a relatively small number of attributes are involved. Under these scenarios, attribute descent has quicker convergence and superior performance compared with existing gradient-free optimization methods. As such, the attribute decent serves as an effective baseline for object-centric content-level domain adaptation.% for a task involving more complex distribution, \ie approximating street scenes, .}

% \textcolor{blue}{It is worth noting that Xue \etal has previously published a paper that validates the effectiveness of our proposed method in the context of a more intricate data distribution, \ie street scenes. This validation was carried out using a supervised loss (\ie task accuracy), which is distinct from the unsupervised loss (\ie domain gap)~\cite{xue2021learning}. }

\begin{figure}[t]
 \begin{minipage}[c]{0.21\textwidth}
    \includegraphics[width=\textwidth]{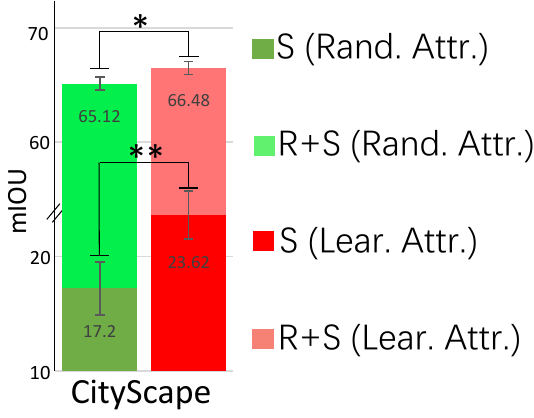}
  \end{minipage}
  % \vspace{-1mm}
  % \hfill
\begin{minipage}[c]{0.27\textwidth}
    \caption{Comparison of training sets synthesized from learned attributes by attribute descent and random attributes, with application to street scene semantic segmentation. We aim to approximate target distribution of street scenes. Similar to Fig.~\ref{figure: stas_analy}, two application scenarios are evaluated: training with synthetic data only (``S'') and training data augmentation (``R+S''). Statistical significance analysis is conducted. }
  \label{fig:scene_attribute_descent}
  \end{minipage}
  % \vspace{-4mm}
\end{figure}

Beyond object-centric tasks, we experimentally show that attribute descent is also useful in the semantic segmentation task where the street scenes have more complex distributions. For this application, we use the 3D assets (\ie SceneX) collected in our previous work \cite{xue2021learning}, where 23 controllable attributes are defined, including scene layout, illumination, \textit{etc}.  % The content-level domain adaptation in the street scene task is notably more challenging. To illustrate, the task of approximating the data distribution of a street scene involves defining 23 distinct controllable attributes, encompassing scene layout attributes, illumination attributes, and more. 
This is significantly more than the 6 attributes defined in the object-centric tasks. Using the Cityscapes dataset~\cite{cordts2016cityscapes} as target and DeepLab-v2 segmentation model~\cite{CP2018Deeplab}, we present quantitative results in Fig.\ref{fig:scene_attribute_descent}. In spite of the more challenging setup, we find that our method still maintains its superiority to random attributes. 
%In contrast, the object-centric task presented in this submission comprises a mere 6 defined controllable attributes. Remarkably, even in the face of such complexity, our findings as reported in Fig. \ref{fig:scene_attribute_descent} indicate that attribute descent still maintains its superiority over random attributes. 
For example, under synthetic only training, attribute descent yields +6.42\% improvement in mean intersection over union (mIoU) over the use of random attributes. That said, inheriting from coordinate descent, attribute descent may have lower running efficiency in a complex scenario \cite{nesterov2012efficiency}, where we speculate that global optimization algorithms like reinforcement learning will be good alternatives.

\section{Conclusion}
This article studies how to improve training data quality from the perspective of reducing the domain gap between synthetic data and real data on the content level. Specifically, we propose an attribute descent algorithm that can automatically edit the source domain (synthetic) image content in a graphic engine to generate training data with a good resemblance to the target domain (real world). % in order to reduce the content gap between the synthetic source images and the real target images. 
We evaluate this method on object-centric tasks, in which the usage of object bounding boxes decreases the number of attributes to be optimized. Fewer attributes of interest allow us to optimize attributes individually using the proposed attribute descent approach. We show that data synthesized from learned attributes improve task accuracy in two application scenarios: training with synthetic data only and augmenting target data with synthetic data. In addition, using viewpoint as an example, we show that attribute descent enables understanding of the dataset content by computing the attribute distribution of given categories or cameras. This article demonstrates the benefit of training data engineering, and in the future, more investigations will be performed to understand training data quality. % Moreover, our experiment reveals some important insights regarding the usage of synthetic data, \emph{e.g.,} style DA brings significant improvement and two stage training is beneficial for joint training. % you used content level earlier, here and one other place a few paragraphs ago, you switched to using the word "appearance" ... ?

% We will continue investigating data simulation and its role in computer vision. Potential directions include more effective objective functions, alternative derivative-free optimization methods, and new recognition training schemes. 
% \footnotetext{}
\section*{Acknowledgement}
This work was supported in part by the ARC Discovery Early Career Researcher Award (DE200101283) and the ARC Discovery Project (DP210102801).

\bibliographystyle{IEEEtran}
\bibliography{content}

% Generated by IEEEtran.bst, version: 1.14 (2015/08/26)
\begin{thebibliography}{10}
\providecommand{\url}[1]{#1}
\csname url@samestyle\endcsname
\providecommand{\newblock}{\relax}
\providecommand{\bibinfo}[2]{#2}
\providecommand{\BIBentrySTDinterwordspacing}{\spaceskip=0pt\relax}
\providecommand{\BIBentryALTinterwordstretchfactor}{4}
\providecommand{\BIBentryALTinterwordspacing}{\spaceskip=\fontdimen2\font plus
\BIBentryALTinterwordstretchfactor\fontdimen3\font minus
  \fontdimen4\font\relax}
\providecommand{\BIBforeignlanguage}[2]{{%
\expandafter\ifx\csname l@#1\endcsname\relax
\typeout{** WARNING: IEEEtran.bst: No hyphenation pattern has been}%
\typeout{** loaded for the language `#1'. Using the pattern for}%
\typeout{** the default language instead.}%
\else
\language=\csname l@#1\endcsname
\fi
#2}}
\providecommand{\BIBdecl}{\relax}
\BIBdecl

\bibitem{richter2016playing}
S.~R. Richter, V.~Vineet, S.~Roth, and V.~Koltun, ``Playing for data: Ground
  truth from computer games,'' in \emph{European Conference on Computer
  Vision}, 2016.

\bibitem{sakaridis2018semantic}
C.~Sakaridis, D.~Dai, and L.~Van~Gool, ``Semantic foggy scene understanding
  with synthetic data,'' \emph{International Journal of Computer Vision}, pp.
  1--20, 2018.

\bibitem{ruiz2019learning}
N.~Ruiz, S.~Schulter, and M.~Chandraker, ``Learning to simulate,'' in
  \emph{Proceedings of the International Conference on Learning
  Representations}, 2019.

\bibitem{tremblay2018training}
J.~Tremblay, A.~Prakash, D.~Acuna, M.~Brophy, V.~Jampani, C.~Anil, T.~To,
  E.~Cameracci, S.~Boochoon, and S.~Birchfield, ``Training deep networks with
  synthetic data: Bridging the reality gap by domain randomization,'' in
  \emph{Proceedings of the IEEE/CVF Conference on Computer Vision and Pattern
  Recognition Workshops}, 2018.

\bibitem{sun2019dissecting}
X.~Sun and L.~Zheng, ``Dissecting person re-identification from the viewpoint
  of viewpoint,'' in \emph{Proceedings of the IEEE/CVF Conference on Computer
  Vision and Pattern Recognition}, 2019.

\bibitem{kar2019meta}
A.~Kar, A.~Prakash, M.-Y. Liu, E.~Cameracci, J.~Yuan, M.~Rusiniak, D.~Acuna,
  A.~Torralba, and S.~Fidler, ``Meta-sim: Learning to generate synthetic
  datasets,'' in \emph{Proceedings of the IEEE International Conference on
  Computer Vision}, 2019.

\bibitem{hoffman2018cycada}
J.~Hoffman, E.~Tzeng, T.~Park, J.-Y. Zhu, P.~Isola, K.~Saenko, A.~Efros, and
  T.~Darrell, ``Cycada: Cycle-consistent adversarial domain adaptation,'' in
  \emph{International Conference on Machine Learning}, 2018.

\bibitem{liu2016deep}
H.~Liu, Y.~Tian, Y.~Yang, L.~Pang, and T.~Huang, ``Deep relative distance
  learning: Tell the difference between similar vehicles,'' in
  \emph{Proceedings of the IEEE/CVF Conference on Computer Vision and Pattern
  Recognition}, 2016.

\bibitem{liu2016large}
X.~Liu, W.~Liu, H.~Ma, and H.~Fu, ``Large-scale vehicle re-identification in
  urban surveillance videos,'' in \emph{The IEEE International Conference on
  Multimedia and Expo}, 2016.

\bibitem{heusel2017gans}
M.~Heusel, H.~Ramsauer, T.~Unterthiner, B.~Nessler, and S.~Hochreiter, ``Gans
  trained by a two time-scale update rule converge to a local nash
  equilibrium,'' in \emph{Advances in neural information processing systems},
  2017.

\bibitem{yao2019simulating}
Y.~Yao, L.~Zheng, X.~Yang, M.~Naphade, and T.~Gedeon, ``Simulating content
  consistent vehicle datasets with attribute descent,'' in \emph{Proceedings of
  the European Conference on Computer Vision}, 2020.

\bibitem{zhu2017unpaired}
J.-Y. Zhu, T.~Park, P.~Isola, and A.~A. Efros, ``Unpaired image-to-image
  translation using cycle-consistent adversarial networks,'' in
  \emph{Proceedings of the IEEE/CVF Conference on Computer Vision and Pattern
  Recognition}, 2017.

\bibitem{shrivastava2017learning}
A.~Shrivastava, T.~Pfister, O.~Tuzel, J.~Susskind, W.~Wang, and R.~Webb,
  ``Learning from simulated and unsupervised images through adversarial
  training,'' in \emph{Proceedings of the IEEE/CVF Conference on Computer
  Vision and Pattern Recognition}, 2017.

\bibitem{deng2018image}
W.~Deng, L.~Zheng, Q.~Ye, G.~Kang, Y.~Yang, and J.~Jiao, ``Image-image domain
  adaptation with preserved self-similarity and domain-dissimilarity for person
  re-identification,'' in \emph{Proceedings of the IEEE/CVF Conference on
  Computer Vision and Pattern Recognition}, 2018.

\bibitem{li2019bidirectional}
Y.~Li, L.~Yuan, and N.~Vasconcelos, ``Bidirectional learning for domain
  adaptation of semantic segmentation,'' in \emph{Proceedings of the IEEE/CVF
  Conference on Computer Vision and Pattern Recognition}, 2019, pp. 6936--6945.

\bibitem{chen2019crdoco}
Y.-C. Chen, Y.-Y. Lin, M.-H. Yang, and J.-B. Huang, ``Crdoco: Pixel-level
  domain transfer with cross-domain consistency,'' in \emph{Proceedings of the
  IEEE/CVF Conference on Computer Vision and Pattern Recognition}, 2019, pp.
  1791--1800.

\bibitem{kouw2016feature}
W.~M. Kouw, L.~J. Van Der~Maaten, J.~H. Krijthe, and M.~Loog, ``Feature-level
  domain adaptation,'' \emph{The Journal of Machine Learning Research},
  vol.~17, no.~1, pp. 5943--5974, 2016.

\bibitem{sun2016return}
B.~Sun, J.~Feng, and K.~Saenko, ``Return of frustratingly easy domain
  adaptation,'' in \emph{Proceedings of the AAAI Conference on Artificial
  Intelligence}, vol.~30, no.~1, 2016.

\bibitem{long2015learning}
M.~Long, Y.~Cao, J.~Wang, and M.~Jordan, ``Learning transferable features with
  deep adaptation networks,'' in \emph{Proceedings of International Conference
  on Machine Learning}, 2015, pp. 97--105.

\bibitem{tzeng2014deep}
E.~Tzeng, J.~Hoffman, N.~Zhang, K.~Saenko, and T.~Darrell, ``Deep domain
  confusion: Maximizing for domain invariance,'' \emph{arXiv preprint
  arXiv:1412.3474}, 2014.

\bibitem{peng2019moment}
X.~Peng, Q.~Bai, X.~Xia, Z.~Huang, K.~Saenko, and B.~Wang, ``Moment matching
  for multi-source domain adaptation,'' in \emph{Proceedings of the IEEE/CVF
  International Conference on Computer Vision}, 2019, pp. 1406--1415.

\bibitem{zhang2018aligning}
Z.~Zhang, M.~Wang, Y.~Huang, and A.~Nehorai, ``Aligning infinite-dimensional
  covariance matrices in reproducing kernel hilbert spaces for domain
  adaptation,'' in \emph{Proceedings of the IEEE/CVF Conference on Computer
  Vision and Pattern Recognition}, 2018, pp. 3437--3445.

\bibitem{zou2020dgnetpp}
Y.~Zou, X.~Yang, Z.~Yu, V.~Kumar, and J.~Kautz, ``Joint disentangling and
  adaptation for cross-domain person re-identification,'' in \emph{European
  Conference on Computer Vision}, 2020.

\bibitem{peng2017visda}
X.~Peng, B.~Usman, N.~Kaushik, J.~Hoffman, D.~Wang, and K.~Saenko, ``Visda: The
  visual domain adaptation challenge,'' \emph{arXiv preprint arXiv:1710.06924},
  2017.

\bibitem{wang2020surpassing}
Y.~Wang, S.~Liao, and L.~Shao, ``Surpassing real-world source training data:
  Random 3d characters for generalizable person re-identification,'' in
  \emph{Proceedings of the 28th ACM International Conference on Multimedia},
  2020, pp. 3422--3430.

\bibitem{zhang2021unrealperson}
T.~Zhang, L.~Xie, L.~Wei, Z.~Zhuang, Y.~Zhang, B.~Li, and Q.~Tian,
  ``Unrealperson: An adaptive pipeline towards costless person
  re-identification,'' in \emph{Proceedings of the IEEE/CVF Conference on
  Computer Vision and Pattern Recognition}, 2021, pp. 11\,506--11\,515.

\bibitem{tang2019pamtri}
Z.~Tang, M.~Naphade, S.~Birchfield, J.~Tremblay, W.~Hodge, R.~Kumar, S.~Wang,
  and X.~Yang, ``Pamtri: Pose-aware multi-task learning for vehicle
  re-identification using highly randomized synthetic data,'' in
  \emph{Proceedings of the IEEE International Conference on Computer Vision},
  2019.

\bibitem{gaidon2016virtual}
A.~Gaidon, Q.~Wang, Y.~Cabon, and E.~Vig, ``Virtual worlds as proxy for
  multi-object tracking analysis,'' in \emph{Proceedings of the IEEE/CVF
  Conference on Computer Vision and Pattern Recognition}, 2016.

\bibitem{xue2021learning}
Z.~Xue, W.~Mao, and L.~Zheng, ``Learning to simulate complex scenes for street
  scene segmentation,'' \emph{IEEE Transactions on Multimedia}, 2021.

\bibitem{kolve2017ai2}
E.~Kolve, R.~Mottaghi, D.~Gordon, Y.~Zhu, A.~Gupta, and A.~Farhadi, ``Ai2-thor:
  An interactive 3d environment for visual ai,'' \emph{arXiv preprint
  arXiv:1712.05474}, 2017.

\bibitem{hou2020multiview}
Y.~Hou, L.~Zheng, and S.~Gould, ``Multiview detection with feature perspective
  transformation,'' in \emph{Proceedings of the European Conference on Computer
  Vision}, 2020.

\bibitem{xiang2020unsupervised}
S.~Xiang, Y.~Fu, G.~You, and T.~Liu, ``Unsupervised domain adaptation through
  synthesis for person re-identification,'' in \emph{Proceedings of the IEEE
  International Conference on Multimedia and Expo}, 2020, pp. 1--6.

\bibitem{hu2019sail}
Y.-T. Hu, H.-S. Chen, K.~Hui, J.-B. Huang, and A.~G. Schwing, ``Sail-vos:
  Semantic amodal instance level video object segmentation-a synthetic dataset
  and baselines,'' in \emph{Proceedings of the IEEE/CVF Conference on Computer
  Vision and Pattern Recognition}, 2019.

\bibitem{wang2019learning}
Q.~Wang, J.~Gao, W.~Lin, and Y.~Yuan, ``Learning from synthetic data for crowd
  counting in the wild,'' in \emph{Proceedings of the IEEE/CVF Conference on
  Computer Vision and Pattern Recognition}, 2019, pp. 8198--8207.

\bibitem{doan2018g2d}
A.-D. Doan, A.~M. Jawaid, T.-T. Do, and T.-J. Chin, ``G2d: from gta to data,''
  \emph{arXiv preprint arXiv:1806.07381}, 2018.

\bibitem{dosovitskiy2017carla}
A.~Dosovitskiy, G.~Ros, F.~Codevilla, A.~Lopez, and V.~Koltun, ``Carla: An open
  urban driving simulator,'' \emph{arXiv preprint arXiv:1711.03938}, 2017.

\bibitem{deitke2020robothor}
M.~Deitke, W.~Han, A.~Herrasti, A.~Kembhavi, E.~Kolve, R.~Mottaghi,
  J.~Salvador, D.~Schwenk, E.~VanderBilt, M.~Wallingford \emph{et~al.},
  ``Robothor: An open simulation-to-real embodied ai platform,'' in
  \emph{Proceedings of the IEEE/CVF Conference on Computer Vision and Pattern
  Recognition}, 2020, pp. 3164--3174.

\bibitem{mueller2017sim4cv}
M.~Mueller, V.~Casser, J.~Lahoud, N.~Smith, and B.~Ghanem, ``Sim4cv: A
  photo-realistic simulator for computer vision applications,''
  \emph{International Journal of Computer Vision}, 08 2017.

\bibitem{tobin2017domain}
J.~Tobin, R.~Fong, A.~Ray, J.~Schneider, W.~Zaremba, and P.~Abbeel, ``Domain
  randomization for transferring deep neural networks from simulation to the
  real world,'' in \emph{2017 IEEE/RSJ international conference on intelligent
  robots and systems (IROS)}.\hskip 1em plus 0.5em minus 0.4em\relax IEEE,
  2017, pp. 23--30.

\bibitem{mozifian2020intervention}
M.~Mozifian, A.~Zhang, J.~Pineau, and D.~Meger, ``Intervention design for
  effective sim2real transfer,'' \emph{arXiv preprint arXiv:2012.02055}, 2020.

\bibitem{jones2020shapeassembly}
R.~K. Jones, T.~Barton, X.~Xu, K.~Wang, E.~Jiang, P.~Guerrero, N.~J. Mitra, and
  D.~Ritchie, ``Shapeassembly: Learning to generate programs for 3d shape
  structure synthesis,'' \emph{ACM Transactions on Graphics (TOG)}, vol.~39,
  no.~6, pp. 1--20, 2020.

\bibitem{yin20213dstylenet}
K.~Yin, J.~Gao, M.~Shugrina, S.~Khamis, and S.~Fidler, ``3dstylenet: Creating
  3d shapes with geometric and texture style variations,'' in \emph{Proceedings
  of the IEEE/CVF International Conference on Computer Vision}, 2021, pp.
  12\,456--12\,465.

\bibitem{shi2019face}
T.~Shi, Y.~Yuan, C.~Fan, Z.~Zou, Z.~Shi, and Y.~Liu, ``Face-to-parameter
  translation for game character auto-creation,'' in \emph{Proceedings of the
  IEEE/CVF International Conference on Computer Vision}, 2019, pp. 161--170.

\bibitem{zhang2020generating}
Y.~Zhang, M.~Hassan, H.~Neumann, M.~J. Black, and S.~Tang, ``Generating 3d
  people in scenes without people,'' in \emph{Proceedings of the IEEE/CVF
  Conference on Computer Vision and Pattern Recognition}, 2020, pp. 6194--6204.

\bibitem{paschalidou2021neural}
D.~Paschalidou, A.~Katharopoulos, A.~Geiger, and S.~Fidler, ``Neural parts:
  Learning expressive 3d shape abstractions with invertible neural networks,''
  in \emph{Proceedings of the IEEE/CVF Conference on Computer Vision and
  Pattern Recognition}, 2021, pp. 3204--3215.

\bibitem{devaranjan2020meta}
J.~Devaranjan, A.~Kar, and S.~Fidler, ``Meta-sim2: Unsupervised learning of
  scene structure for synthetic data generation,'' in \emph{European Conference
  on Computer Vision}.\hskip 1em plus 0.5em minus 0.4em\relax Springer, 2020,
  pp. 715--733.

\bibitem{deng2009imagenet}
J.~Deng, W.~Dong, R.~Socher, L.-J. Li, K.~Li, and L.~Fei-Fei, ``Imagenet: A
  large-scale hierarchical image database,'' in \emph{Proceedings of the
  IEEE/CVF Conference on Computer Vision and Pattern Recognition}, 2009.

\bibitem{lin2014microsoft}
T.-Y. Lin, M.~Maire, S.~Belongie, J.~Hays, P.~Perona, D.~Ramanan,
  P.~Doll{\'a}r, and C.~L. Zitnick, ``Microsoft coco: Common objects in
  context,'' in \emph{European conference on computer vision}.\hskip 1em plus
  0.5em minus 0.4em\relax Springer, 2014, pp. 740--755.

\bibitem{he2016deep}
K.~He, X.~Zhang, S.~Ren, and J.~Sun, ``Deep residual learning for image
  recognition,'' in \emph{Proceedings of the IEEE/CVF Conference on Computer
  Vision and Pattern Recognition}, 2016, pp. 770--778.

\bibitem{simonyan2014very}
K.~Simonyan and A.~Zisserman, ``Very deep convolutional networks for
  large-scale image recognition,'' \emph{arXiv preprint arXiv:1409.1556}, 2014.

\bibitem{zheng2015scalable}
L.~Zheng, L.~Shen, L.~Tian, S.~Wang, J.~Wang, and Q.~Tian, ``Scalable person
  re-identification: A benchmark,'' in \emph{Proceedings of the IEEE
  International Conference on Computer Vision}, 2015.

\bibitem{ristani2016MTMC}
E.~Ristani, F.~Solera, R.~Zou, R.~Cucchiara, and C.~Tomasi, ``Performance
  measures and a data set for multi-target, multi-camera tracking,'' in
  \emph{European Conference on Computer Vision workshop on Benchmarking
  Multi-Target Tracking}, 2016.

\bibitem{khorramshahi2019dual}
P.~Khorramshahi, A.~Kumar, N.~Peri, S.~S. Rambhatla, J.-C. Chen, and
  R.~Chellappa, ``A dual path modelwith adaptive attention for vehicle
  re-identification,'' in \emph{Proceedings of the IEEE International
  Conference on Computer Vision}, 2019.

\bibitem{wang2017orientation}
Z.~Wang, L.~Tang, X.~Liu, Z.~Yao, S.~Yi, J.~Shao, J.~Yan, S.~Wang, H.~Li, and
  X.~Wang, ``Orientation invariant feature embedding and spatial temporal
  regularization for vehicle re-identification,'' in \emph{Proceedings of the
  IEEE International Conference on Computer Vision}, 2017.

\bibitem{zhou2018aware}
Y.~Zhou and L.~Shao, ``Aware attentive multi-view inference for vehicle
  re-identification,'' in \emph{Proceedings of the IEEE/CVF Conference on
  Computer Vision and Pattern Recognition}, 2018.

\bibitem{torralba2011unbiased}
A.~Torralba and A.~A. Efros, ``Unbiased look at dataset bias,'' in
  \emph{Proceedings of the European Conference on Computer Vision}, 2011.

\bibitem{kuznetsova2020open}
A.~Kuznetsova, H.~Rom, N.~Alldrin, J.~Uijlings, I.~Krasin, J.~Pont-Tuset,
  S.~Kamali, S.~Popov, M.~Malloci, A.~Kolesnikov \emph{et~al.}, ``The open
  images dataset v4,'' \emph{International Journal of Computer Vision}, vol.
  128, no.~7, pp. 1956--1981, 2020.

\bibitem{barbu2019objectnet}
A.~Barbu, D.~Mayo, J.~Alverio, W.~Luo, C.~Wang, D.~Gutfreund, J.~Tenenbaum, and
  B.~Katz, ``Objectnet: A large-scale bias-controlled dataset for pushing the
  limits of object recognition models,'' in \emph{Advances in Neural
  Information Processing Systems}, 2019.

\bibitem{van2008visualizing}
L.~Van~der Maaten and G.~Hinton, ``Visualizing data using t-sne.''
  \emph{Journal of machine learning research}, vol.~9, no.~11, 2008.

\bibitem{zhong2018camstyle}
Z.~Zhong, L.~Zheng, Z.~Zheng, S.~Li, and Y.~Yang, ``Camstyle: A novel data
  augmentation method for person re-identification,'' \emph{IEEE Transactions
  on Image Processing}, vol.~28, no.~3, pp. 1176--1190, 2018.

\bibitem{chang2015shapenet}
A.~X. Chang, T.~Funkhouser, L.~Guibas, P.~Hanrahan, Q.~Huang, Z.~Li,
  S.~Savarese, M.~Savva, S.~Song, H.~Su \emph{et~al.}, ``Shapenet: An
  information-rich 3d model repository,'' \emph{arXiv preprint
  arXiv:1512.03012}, 2015.

\bibitem{miller1995wordnet}
G.~A. Miller, ``Wordnet: a lexical database for english,'' \emph{Communications
  of the ACM}, vol.~38, no.~11, pp. 39--41, 1995.

\bibitem{juliani2018unity}
A.~Juliani, V.-P. Berges, E.~Vckay, Y.~Gao, H.~Henry, M.~Mattar, and D.~Lange,
  ``Unity: A general platform for intelligent agents,'' \emph{arXiv preprint
  arXiv:1809.02627}, 2018.

\bibitem{tang2019cityflow}
Z.~Tang, M.~Naphade, M.-Y. Liu, X.~Yang, S.~Birchfield, S.~Wang, R.~Kumar,
  D.~Anastasiu, and J.-N. Hwang, ``Cityflow: A city-scale benchmark for
  multi-target multi-camera vehicle tracking and re-identification,'' in
  \emph{Proceedings of the IEEE/CVF Conference on Computer Vision and Pattern
  Recognition}, 2019.

\bibitem{ruiz2018learning}
N.~Ruiz, S.~Schulter, and M.~Chandraker, ``Learning to simulate,'' in
  \emph{Proceedings of the International Conference on Learning
  Representations}, 2018.

\bibitem{szegedy2016rethinking}
C.~Szegedy, V.~Vanhoucke, S.~Ioffe, J.~Shlens, and Z.~Wojna, ``Rethinking the
  inception architecture for computer vision,'' in \emph{Proceedings of the
  IEEE/CVF Conference on Computer Vision and Pattern Recognition}, 2016.

\bibitem{kim2017learning}
T.~Kim, M.~Cha, H.~Kim, J.~K. Lee, and J.~Kim, ``Learning to discover
  cross-domain relations with generative adversarial networks,'' in
  \emph{International Conference on Machine Learning}, 2017.

\bibitem{girshick2015fast}
R.~Girshick, ``Fast r-cnn,'' in \emph{Proceedings of the IEEE international
  conference on computer vision}, 2015, pp. 1440--1448.

\bibitem{wright2015coordinate}
S.~J. Wright, ``Coordinate descent algorithms,'' \emph{Mathematical
  Programming}, vol. 151, no.~1, pp. 3--34, 2015.

\bibitem{deng2018similarity}
W.~Deng, L.~Zheng, Q.~Ye, Y.~Yang, and J.~Jiao, ``Similarity-preserving
  image-image domain adaptation for person re-identification,'' \emph{arXiv
  preprint arXiv:1811.10551}, 2018.

\bibitem{zhong2019camstyle}
Z.~Zhong, L.~Zheng, Z.~Zheng, S.~Li, and Y.~Yang, ``Camstyle: A novel data
  augmentation method for person re-identification,'' \emph{IEEE Transactions
  on Image Processing}, vol.~28, no.~3, pp. 1176--1190, 2019.

\bibitem{zheng2019vehiclenet}
Z.~Zheng, T.~Ruan, Y.~Wei, and Y.~Yang, ``Vehiclenet: Learning robust feature
  representation for vehicle re-identification,'' in \emph{Proceedings of the
  IEEE/CVF Conference on Computer Vision and Pattern Recognition Workshops},
  2019.

\bibitem{tzeng2017adversarial}
E.~Tzeng, J.~Hoffman, K.~Saenko, and T.~Darrell, ``Adversarial discriminative
  domain adaptation,'' in \emph{Proceedings of the IEEE International
  Conference on Computer Vision}, 2017.

\bibitem{liang2020we}
J.~Liang, D.~Hu, and J.~Feng, ``Do we really need to access the source data?
  source hypothesis transfer for unsupervised domain adaptation,'' in
  \emph{International Conference on Machine Learning}.\hskip 1em plus 0.5em
  minus 0.4em\relax PMLR, 2020, pp. 6028--6039.

\bibitem{zheng2016mars}
L.~Zheng, Z.~Bie, Y.~Sun, J.~Wang, C.~Su, S.~Wang, and Q.~Tian, ``Mars: A video
  benchmark for large-scale person re-identification,'' in \emph{European
  Conference on Computer Vision}, 2016.

\bibitem{sun2018beyond}
Y.~Sun, L.~Zheng, Y.~Yang, Q.~Tian, and S.~Wang, ``Beyond part models: Person
  retrieval with refined part pooling (and a strong convolutional baseline),''
  in \emph{European Conference on Computer Vision}, 2018.

\bibitem{zhuang2020rethinking}
Z.~Zhuang, L.~Wei, L.~Xie, T.~Zhang, H.~Zhang, H.~Wu, H.~Ai, and Q.~Tian,
  ``Rethinking the distribution gap of person re-identification with
  camera-based batch normalization,'' in \emph{European Conference on Computer
  Vision}.\hskip 1em plus 0.5em minus 0.4em\relax Springer, 2020, pp. 140--157.

\bibitem{he2021transreid}
S.~He, H.~Luo, P.~Wang, F.~Wang, H.~Li, and W.~Jiang, ``Transreid:
  Transformer-based object re-identification,'' in \emph{Proceedings of the
  IEEE International Conference on Computer Vision}, 2021.

\bibitem{liu2018ram}
X.~Liu, S.~Zhang, Q.~Huang, and W.~Gao, ``Ram: a region-aware deep model for
  vehicle re-identification,'' in \emph{The IEEE International Conference on
  Multimedia and Expo}, 2018.

\bibitem{bai2018group}
Y.~Bai, Y.~Lou, F.~Gao, S.~Wang, Y.~Wu, and L.-Y. Duan, ``Group-sensitive
  triplet embedding for vehicle reidentification,'' \emph{IEEE Transactions on
  Multimedia}, vol.~20, no.~9, pp. 2385--2399, 2018.

\bibitem{chu2019vehicle}
R.~Chu, Y.~Sun, Y.~Li, Z.~Liu, C.~Zhang, and Y.~Wei, ``Vehicle
  re-identification with viewpoint-aware metric learning,'' in
  \emph{Proceedings of the IEEE International Conference on Computer Vision},
  2019.

\bibitem{luo2019bag}
H.~Luo, Y.~Gu, X.~Liao, S.~Lai, and W.~Jiang, ``Bag of tricks and a strong
  baseline for deep person re-identification,'' in \emph{Proceedings of the
  IEEE/CVF Conference on Computer Vision and Pattern Recognition Workshops},
  2019.

\bibitem{bergstra2012random}
J.~Bergstra and Y.~Bengio, ``Random search for hyper-parameter optimization,''
  \emph{Journal of machine learning research}, vol.~13, no. Feb, pp. 281--305,
  2012.

\bibitem{xie2017genetic}
L.~Xie and A.~Yuille, ``Genetic cnn,'' in \emph{Proceedings of the IEEE
  international conference on computer vision}, 2017, pp. 1379--1388.

\bibitem{shahriari2015taking}
B.~Shahriari, K.~Swersky, Z.~Wang, R.~P. Adams, and N.~De~Freitas, ``Taking the
  human out of the loop: A review of bayesian optimization,'' \emph{Proceedings
  of the IEEE}, vol. 104, no.~1, pp. 148--175, 2015.

\bibitem{deng2021labels}
W.~Deng and L.~Zheng, ``Are labels always necessary for classifier accuracy
  evaluation?'' in \emph{Proceedings of the IEEE/CVF Conference on Computer
  Vision and Pattern Recognition}, 2021, pp. 15\,069--15\,078.

\bibitem{deng2021does}
W.~Deng, S.~Gould, and L.~Zheng, ``What does rotation prediction tell us about
  classifier accuracy under varying testing environments?'' in
  \emph{International Conference on Machine Learning}, 2021.

\bibitem{cordts2016cityscapes}
M.~Cordts, M.~Omran, S.~Ramos, T.~Rehfeld, M.~Enzweiler, R.~Benenson,
  U.~Franke, S.~Roth, and B.~Schiele, ``The cityscapes dataset for semantic
  urban scene understanding,'' in \emph{Proceedings of the IEEE Computer Vision
  and Pattern Recognition}, 2016, pp. 3213--3223.

\bibitem{CP2018Deeplab}
L.-C. Chen, G.~Papandreou, I.~Kokkinos, K.~Murphy, and A.~L. Yuille, ``Deeplab:
  Semantic image segmentation with deep convolutional nets, atrous convolution,
  and fully connected crfs,'' \emph{IEEE Transactions on Pattern Analysis and
  Machine Intelligence}, 2018.

\bibitem{nesterov2012efficiency}
Y.~Nesterov, ``Efficiency of coordinate descent methods on huge-scale
  optimization problems,'' \emph{SIAM Journal on Optimization}, vol.~22, no.~2,
  pp. 341--362, 2012.

\end{thebibliography}

 \begin{IEEEbiography}
[{\includegraphics[width=1in,height=1.5in,clip,keepaspectratio]{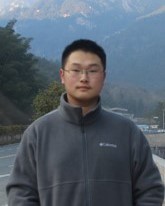}}]{Yue Yao} is a Research Officer at the School of Computing, Australian National University. Before that, he studied as a Ph.D. student at the Australian National University. He completed the degree of Bachelor of Advanced Computing at the Australian National University in 2018. His research interests include training set optimization, data synthesis and brain-computer interface.
\end{IEEEbiography}

\begin{IEEEbiography}
[{\includegraphics[width=1in,height=1.5in,clip,keepaspectratio]{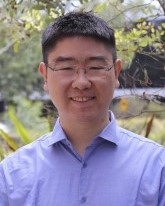}}]{Liang Zheng} is an Associate Professor at the School of Computing, Australian National University. He received the Ph.D. degree in Electronic Engineering from Tsinghua University, China, in 2015, and the B.E. degree in Life Science from Tsinghua University, China, in 2010. He was a postdoc researcher at the Center for Artificial Intelligence, University of Technology Sydney, Australia (2016 - 2018). His research interests include object re-identification, domain adaptation, deep learning and data synthesis.
\end{IEEEbiography}
 %\vspace{-15mm}
 
\begin{IEEEbiography}
[{\includegraphics[width=1in,height=1.5in,clip,keepaspectratio]{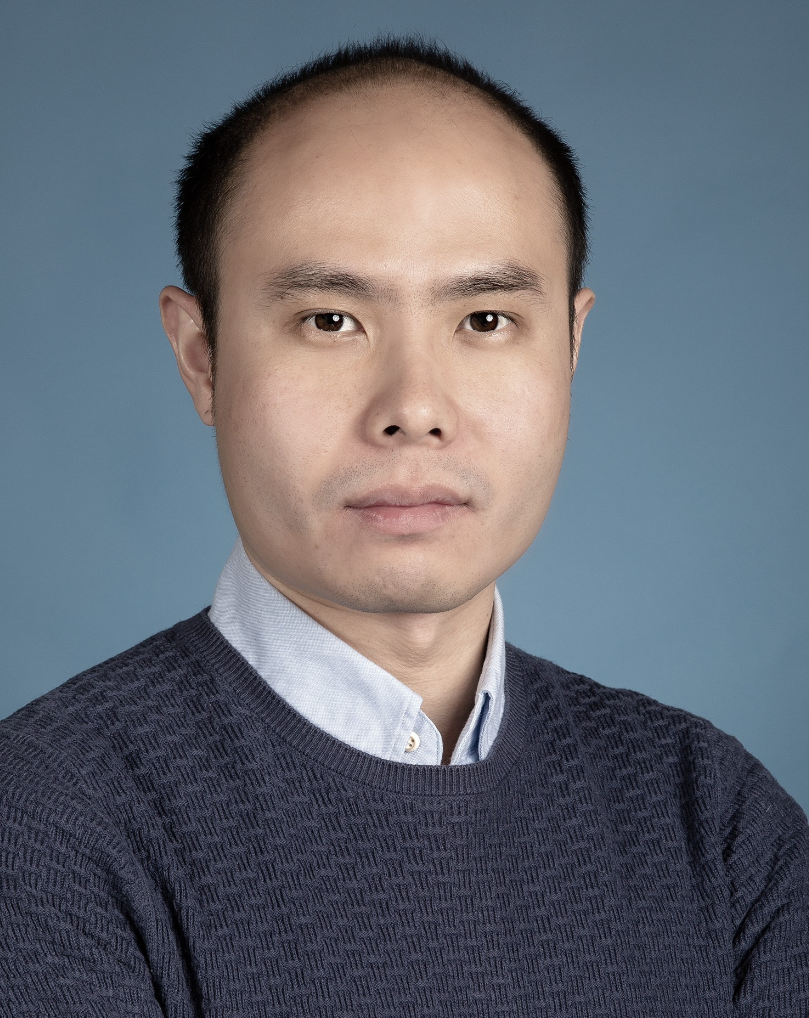}}]{Xiaodong Yang} is the Head of Machine Learning at QCraft. Previously, he was a Senior Research Scientist at NVIDIA Research. He received the B.S. degree from Huazhong University of Science and Technology, China, in 2009, and the Ph.D. degree from City University of New York, USA, in 2015. His research interests include autonomous driving, image and video understanding, activity and gesture recognition, facial analytics, object re-identification, deep generative modeling, etc. 
\end{IEEEbiography}
 %\vspace{-15mm}
 
 \begin{IEEEbiography}
[{\includegraphics[width=1in,height=1.5in,clip,keepaspectratio]{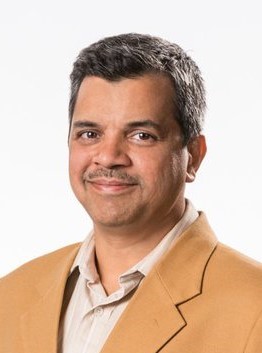}}]{Milind Napthade} is the Senior Vice President in Capital One. Previously he led the technology and innovation strategy and engineering execution for NVIDIA’s Metropolis platform. He previously served as the program director in the Smarter City Services Group at the IBM T.J. Watson Research Center
in Hawthorne, New York, and leader of the Smarter Sustainable Dubuque living lab. Naphade received a Ph.D. in electrical engineering from the University of Illinois at Urbana-Champaign. 
% He is a senior member of IEEE and a member of the IEEE Circuits and Systems Society’s Multimedia Systems and Applications Technical Committee. 
% Contact him at naphade@us.ibm.com.
\end{IEEEbiography}
 %\vspace{-15mm}
 
 \begin{IEEEbiography}
[{\includegraphics[width=1in,height=1.5in,clip,keepaspectratio]{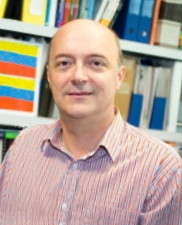}}]{Tom Gedeon} is the Human-Centric Advancements Chair in AI and Head of the Human-Centric Advancements group in Artificial Intelligence at Curtin University. He is an international Research Professor at Obuda University in Hungary. He is an Honorary Professor at the Australian National University, where he was formerly Deputy Dean and Head of Computer Science. His B.Sc and Ph.D. are from the University of Western Australia, and Grad Dip Management from UNSW. He is twice a former President of the Asia-Pacific Neural Network Assembly, and former President of the Computing Research and Education Association of Australasia. He is an associate editor of the IEEE Transactions on Fuzzy Systems, and the INNS/Elsevier journal Neural Networks. His research interests are in responsive AI and responsible AI. 
\end{IEEEbiography}
%\vspace{-15mm}
 
\vfill
% that's all folks
\end{document}